\documentclass[10pt,journal,compsoc]{IEEEtran}

\usepackage[utf8]{inputenc} %
\usepackage[T1]{fontenc}    %
\usepackage{hyperref}       %
\usepackage{url, eucal}            %
\usepackage{booktabs}       %
\usepackage{amsfonts}       %
\usepackage{nicefrac}       %
\usepackage{microtype}      %

\usepackage[table]{xcolor}
\usepackage{amsmath}
\usepackage{caption, subcaption, multirow, overpic, textpos}
\usepackage{booktabs}
\usepackage{xspace}
\usepackage{balance}
\usepackage{makecell}
\usepackage{array}
\usepackage{numprint}
\usepackage{sidecap}
\usepackage{float}
\usepackage{algorithm}

\usepackage{algorithmic}
\usepackage{wrapfig}
\usepackage{gensymb}
\usepackage[final]{changes}

\usepackage[capitalize]{cleveref}
\crefname{section}{Sec.}{Secs.}
\Crefname{section}{Section}{Sections}
\Crefname{table}{Table}{Tables}
\crefname{table}{Table}{Tabs.}

\newcolumntype{B}[1]{>{\centering\arraybackslash}b{#1}}

\definecolor{mycolor}{gray}{.69}

\definechangesauthor[name={Reviewer Response 1}, color=blue]{R1}
\definechangesauthor[name={Reviewer Response 2}, color=green]{R2}
\definechangesauthor[name={Reviewer Response 3}, color=orange]{R3}
\definechangesauthor[name={Reviewer Response}, color=green!70!gray]{PR}

\hypersetup{
    colorlinks=true,
    breaklinks=true,
    urlcolor=blue,
    linkcolor=blue,
    bookmarksopen=false,
    filecolor=black,
    citecolor=mycolor,
    linkbordercolor=blue
}

\title{
Diffusion Models are Efficient Data Generators for Human Mesh Recovery
}

\author{Yongtao Ge,
Wenjia Wang,
Yongfan Chen,
Fanzhou Wang,
Lei Yang,
Hao Chen,
Chunhua Shen

\IEEEcompsocitemizethanks{
    \IEEEcompsocthanksitem { \it Accepted to IEEE Trans. Pattern Analysis and Machine Intelligence, Oct. 2025. This version includes the supplementary material, and can be slightly different from the final version.}
    \IEEEcompsocthanksitem Yongtao Ge is with The University of Adelaide. This work was done when he was visiting Zhejiang University.
       \IEEEcompsocthanksitem Chunhua Shen  is
       with Zhejiang
    University of Technology.
    \IEEEcompsocthanksitem Yongtao Ge, Yongfan Chen, Hao Chen, and Chunhua Shen are with Zhejiang University.
    \IEEEcompsocthanksitem Wenjia Wang is with The University of Hong Kong.
    \IEEEcompsocthanksitem Fanzhou Wang and Lei Yang are with SenseTime.
    \IEEEcompsocthanksitem The corresponding author is Chunhua Shen: chunhuashen@zju.edu.cn
}%
}
\begin{document}

\newcommand{\predTheta}{\hat{\mathbf{\Theta}}}
\newcommand{\predShape}{\hat{\mathbf{\beta}}}
\newcommand{\predPose}{\hat{\mathbf{\theta}}}
	
\newcommand{\gtTheta}{\mathbf{\Theta}}
\newcommand{\gtShape}{\mathbf{\beta}}
\newcommand{\gtPose}{\mathbf{\theta}}

\newcommand{\motionDisc}{\mathcal{D}_M}	
\newcommand{\generator}{\mathcal{G}}
\newcommand{\smpl}{\mathcal{M}}	

\newcommand{\focalx}{f_x}
\newcommand{\focaly}{f_y}	
\newcommand{\centerx}{o_x}
\newcommand{\centery}{o_y}
\newcommand{\campos}{C}
\newcommand{\camtransl}{t^c}
\newcommand{\camrot}{R^c}
\newcommand{\imgwidth}{w}
\newcommand{\imgheight}{h}
\newcommand{\cropwidth}{w_{bbox}}
\newcommand{\cropheight}{h_{bbox}}
\newcommand{\cropcenterx}{c_x}
\newcommand{\cropcentery}{c_y}
\newcommand{\predjoints}[1]{\hat{\mathcal{J}}_{\mathit{#1D}}}
\newcommand{\gtjoints}[1]{\mathcal{J}_{\mathit{#1D}}}

\newcommand{\bodytransl}{t^b}
\newcommand{\bodyori}{R^b}

\newcommand{\pampjpe}{PA-MPJPE\xspace}
\newcommand{\wmpjpe}{W-MPJPE\xspace}
\newcommand{\cmpjpe}{C-MPJPE\xspace}
\newcommand{\mpjpe}{MPJPE\xspace}
\newcommand{\wpve}{W-PVE\xspace}
\newcommand{\cpve}{C-PVE\xspace}
\newcommand{\pve}{PVE\xspace}

\newcommand{\iwcam}{IWP-cam\xspace}
\newcommand{\camcalib}{CamCalib\xspace}
\newcommand{\pitch}{\alpha}
\newcommand{\roll}{\phi}
\newcommand{\yaw}{\psi}
\newcommand{\vfov}{\upsilon}
\newcommand{\softltwo}{Softargmax-$\mathcal{L}_{2}$\xspace}
\newcommand{\softbiasedltwo}{Softargmax-biased-$\mathcal{L}_{2}$\xspace}
\newcommand{\ltwo}{$\mathcal{L}_2$\xspace}
\newcommand{\smplify}{SMPLify-X-cam\xspace}

\newcommand{\methodname}{SPEC\xspace} %
\newcommand{\figref}[1]{Fig.~\ref{#1}}

\newcommand{\mpi}{\texttt{MPI-INF-3DHP}\xspace}
\newcommand{\mpii}{\texttt{MPII}\xspace}
\newcommand{\lspet}{\texttt{LSPET}\xspace}
\newcommand{\hthreesixm}{\texttt{Human3.6M}\xspace}
\newcommand{\threedpw}{\texttt{3DPW}\xspace}
\newcommand{\threedpwocc}{\texttt{3DPW-OCC}\xspace}
\newcommand{\coco}{\texttt{COCO}\xspace}
\newcommand{\cocoeft}{\texttt{COCO-EFT}\xspace}
\newcommand{\ooh}{\texttt{3DOH}\xspace}

\newcommand{\agoracam}{\methodname-SYN\xspace}
\newcommand{\mtpcam}{\methodname-MTP\xspace}

\newcommand{\smplifyxc}{SMPLify-XC\xspace}

\newcommand{\supmat}{Sup.~Mat.\xspace}

\newcommand{\real}{\mathbb{R}}

\def\etal{\emph{et al}.}
\def\eg{\emph{e.g}.}
\def\ie{\emph{i.e}.}

\newcommand{\bs}[1]{\boldsymbol{#1}}
\newcommand{\vecb}[1]{{#1}}

\definecolor{myyellow}{rgb}{0.8,0.8,0}
\definecolor{mygreen}{rgb}{0,0.8,0}
\definecolor{myred}{rgb}{0.8,0,0}

\newcommand{\wcon}{W_{\scalebox{0.7}{\textnormal{cn}}}}
\newcommand{\del}{\textnormal{zr}}

\newcommand{\myparagraph}[1]{{\noindent\bf #1}}
\newcommand{\myrowcolor}{\rowcolor[gray]{0.925}}
\newcommand{\gyt}[1]{{\color{blue}#1}}

\newcolumntype{L}[1]{>{\raggedright\arraybackslash}p{#1}}
\newcolumntype{R}[1]{>{\raggedleft\arraybackslash}p{#1}}
\newcolumntype{M}[1]{>{\centering\arraybackslash}p{#1}}

\def\Ours{{{{ \textbf{HumanWild}}}}\xspace}

\IEEEtitleabstractindextext{

\begin{abstract}

    Despite remarkable progress having been made on the problem of 3D human pose and shape estimation (HPS), current state-of-the-art methods rely heavily on either confined indoor mocap datasets or datasets generated by a rendering engine using computer graphics (CG).
    Both categories of datasets exhibit inadequacies in furnishing adequate human identities and authentic in-the-wild background scenes, which are crucial for accurately simulating real-world distributions.
    In this work, we show that synthetic data created by generative models is complementary to CG-rendered data for achieving remarkable generalization performance on diverse real-world scenes. 
    We propose an effective data generation pipeline based on recent diffusion models, termed \textbf{\Ours}, which can effortlessly generate human images and corresponding 3D mesh annotations. 
    Specifically, we first collect a large-scale human-centric dataset with comprehensive annotations, \eg, text captions, the depth map, and surface normal images. 
    To generate a wide variety of human images with initial labels, we train a customized, multi-condition ControlNet model. The key to this process is using a 3D parametric model, \eg, SMPL-X, to easily create precise 2D keypoints, depth maps, and surface normal images by rendering the 3D mesh \added[id=PR]{with specific camera parameters}.
    As there exists inevitable noise in the initial labels, we apply an off-the-shelf foundation segmentation model to filter negative data samples, and a 2D vertex estimator to rectify the SMPL-X parameters by using SMPLify. 
    Our data generation pipeline is both flexible and customizable, making it adaptable to various real-world tasks, such as human interaction in complex scenes and humans captured by wide-angle lenses. By relying solely on generative models, we can produce large-scale, in-the-wild human images with high-quality annotations, significantly reducing the need for manual image collection and annotation. The generated dataset encompasses a wide range of viewpoints, environments, and human identities, ensuring its versatility across different scenarios.
    To verify the effectiveness of the generated data, we perform comprehensive data ablation experiments by performing data ablation studies on top of the both generated data and existing datasets, and evaluating on a wide range of HPS benchmarks. We hope 
    that 
    our work could pave the way for scaling up 3D human recovery to in-the-wild scenes.

\end{abstract}


}

\maketitle

\section{Introduction}
\label{sec:intro}

Estimating human pose and shape (HPS)~\cite{hmr, meshgraphormer, hybrik, cliff,tore} from a single RGB image is a core challenge in computer vision and has many applications in robotics~\cite{case,unihsi, wang2023learning}, computer graphics~\cite{emdm, tlcontrol}, and digital content creation~\cite{Disentangled}. 
Current HPS estimation methods require well-annotated datasets to achieve good performance.
Unfortunately, collecting large-scale, versatile 3D human body data is time-consuming and expensive. 

 Contemporary methodologies for acquiring precise 3D human body data predominantly employ two primary pipelines. The first pipeline involves indoor motion capture (mocap) systems, including both marker-based and vision-based approaches, which are employed by many existing datasets~\cite{h36m, moyo, humman, mpi-inf3dhp} to capture human body attributes. However, this mocap pipeline faces two primary limitations. \added[id=PR]{First, mocap systems are intricate and costly, and their complex operation demands specialized expertise. Second, the data captured is not very diverse. Datasets usually feature a limited number of actors in a lab or studio, which restricts the collection of large-scale human data in more varied environments.}
 To diversify the data sources, some researchers also capture 3D pseudo ground truth by estimating from 2D clues or using additional sensors. SMPLify~\cite{smplify} proposed to fit the parameters of a 3D human model to the location of 2D keypoints. EFT~\cite{eft} introduced the Exemplar Fine-Tuning strategy by overfitting a pre-trained 3D pose regressor with 2D keypoint reprojection loss, taking the final output of the regressor as pseudo labels. However, these methodologies face challenges in accurately generating camera parameters and body parameters, leading to sub-optimal performance on in-the-wild 3D human pose and shape estimation.

\begin{figure*}[ht]
    \centering
    \includegraphics[width=0.99\linewidth]{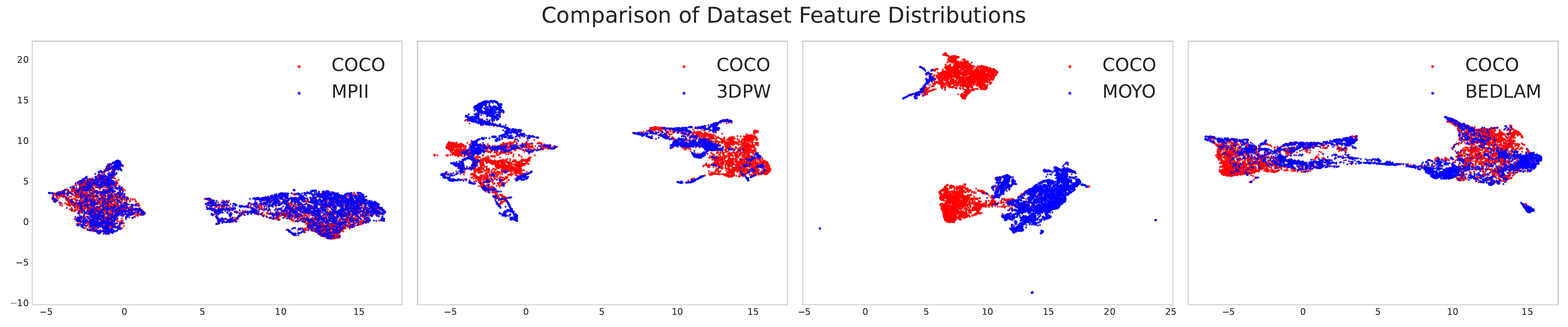}
    \caption{Dataset appearance distributions of synthesized datasets and in-the-wild real-world datasets.}    
    \label{fig:data_distribution}
\end{figure*}

An alternative approach is to generate 3D human datasets using computer-generated (CG) rendering, as seen in works like GTA-Human~\cite{gta-human}, BEDLAM~\cite{bedlam} and SynBody~\cite{synbody}. However, this method has two main challenges. First, there are significant costs involved in obtaining high-quality 3D assets, such as avatars and scene elements, as well as the need for specialized skills in 3D rendering. Second, making synthetic humans and backgrounds look realistic is a major hurdle. There's often a noticeable visual difference between the rendered images and real-world photographs. \added[id=PR]{While Black et al. \cite{bedlam} show the potential of high-quality synthetic data to replace real images, they also highlight important limitations. Their work demonstrates that CLIFF \cite{cliff}, when trained only on synthetic data, can outperform models trained on real datasets. Conversely, they report that training HPS models from scratch with synthetic data, without leveraging pre-training on real-world 2D keypoints datasets (like COCO~\cite{coco}), leads to inferior results. The study further reveals the significant positive impact of realistic clothing simulation compared to more basic textured bodies. This suggests that while synthetic data is a powerful tool, the degree of realism in the data is crucial for advancing 3D human pose and shape estimation.}

\added[id=R1]{While 2D keypoint detectors generalize well when trained on large in-the-wild datasets like COCO, significant gaps exist with 3D datasets. To investigate these gaps, we visualized the feature distributions of several key datasets. We extracted features from cropped human images using DINO-v2 and used UMAP for dimensionality reduction, as shown in \cref{fig:data_distribution}.
Our analysis compares the real-world 2D datasets COCO~\cite{coco} and MPII~\cite{mpii} against three distinct 3D datasets: the real-world outdoor 3DPW~\cite{3dpw}, the indoor Mocap MOYO~\cite{moyo}, and the synthetically rendered BEDLAM~\cite{bedlam}.
The visualization reveals three key insights: (1). The feature distributions of the two real-world 2D datasets, COCO and MPII, are nearly identical. (2). In contrast, both the outdoor 3DPW and indoor MOYO datasets show distributions that diverge significantly from COCO. (3). The CG-rendered BEDLAM dataset is closer to COCO’s distribution than the other 3D datasets, though a noticeable gap remains.
}

\added[id=PR]{This work introduces an automatic and scalable pipeline for generating synthetic data for 3D human pose and shape estimation. Our method uses generative models to create data with more realistic human distributions, overcoming a key limitation of current CG-rendered datasets, that diverse, high-quality CG assets do not scale easily. The intuition is that modern large generative models, \eg, Stable Diffusion XL~\cite{sdxl} and FLUX~\cite{flux2024} are trained on billion-scale text-image pairs. The extensive training regime has shown its efficacy in faithfully simulating real-world distributions.}
The challenge of the pipeline lies in the sampling of diverse data pairs from the prior distribution inherent to generative models.
As analyzed in GTA-Human~\cite{gta-human}, ensuring diversity in camera angles, poses, body shapes, and human appearances within generated human images is critical to accurately simulate real-world human distributions. This diversity is essential for improving the performance of human pose and shape (HPS) tasks by providing more representative and varied training data, especially in-the-wild data.
In contrast, ensuring precise alignment between human images and generated annotations is another critical aspect. This alignment is indispensable to facilitate the effective training of downstream tasks.
DiffusionHPC~\cite{diffusionHPC} offers a straightforward solution by providing the model with diverse text prompts and using pretrained 3D human pose estimation (HPS) models to generate pseudo-labels. However, relying solely on text prompts lacks the granularity needed to precisely control important factors like pose, body shape, and spatial positioning. As a result, the generated 3D pseudo-labels often contain substantial noise, reducing their accuracy.

In contrast, we propose a generative method that leverages precise 3D annotations as conditions. We begin by sampling the SMPL-X parameters from large-scale human motion capture datasets such as AMASS~\cite{amass}. 
Then, we utilize a random camera setup to render the human mesh into three spatial maps, \ie, a keypoint heatmap, a depth map, and a surface normal map. These spatial maps are added as extra input conditions to the generative models, as they provide valuable information about human body’s pose, depth, and surface orientation, enhancing the model's ability to generate more accurate outputs.
Finally, we feed text prompts and these perceptual maps to a multi-condition ControlNet~\cite{controlnet} for generating human images.

Notably, we observe existing general-purpose controllable image generators, \eg, ControlNet~\cite{controlnet}, struggle to generate accurate human images even though precise spatial conditioning controls are provided. This is mainly due to the lack of high-quality training samples.
To address this, we first create a large-scale human-centric dataset, each with detailed annotations including text descriptions, human poses, depth maps, and surface normals. The full data curation pipeline is outlined in~\cref{sec:human_dataset}. We then fine-tune a multi-condition ControlNet using this dataset to improve its performance in generating human images under spatial constraints.

Thanks to the high-quality fine-tuning data, most data samples generated by the multi-condition ControlNet demonstrate satisfactory performance. However, empirical analysis revealed instances of label noise in the initial training pairs. These issues include cases where the generated human figure and input conditions appear as mirrored pairs \added[id=R3]{(see ~\cref{fig:mirror_pair})}, or where the head orientation in the image differs from the input SMPL-X parameters (see failure cases in Supp. Mat.). To address this, we use a pre-trained segmentation model, SAM~\cite{sam}, to predict the human mask in the generated images. We then calculate the intersection-over-union (mIoU) between the ground-truth human mask and the predicted one, to filter out data samples with an IoU below a certain threshold. Finally, we apply a well-trained 2D dense surface estimator to refine the SMPL-X parameters using the SMPLify method~\cite{smplify}, further improving the alignment between the input and generated outputs.

With the aforementioned pipeline, we can finally generate a large-scale 3D human dataset in the wild, with high-quality human appearance and 3D annotations. As shown in~\cref{tab:dataset_comparision}, both mocap-based and CG-rendered datasets are limited in providing diverse human identities and in-the-wild scenes. Compared to previous datasets, our pipeline can generate diverse human identities, human interactions, and various in-the-wild scenes. Notably, the pipeline is much cheaper than both mocap-based and CG-based counterparts and is scalable to generate 3D human datasets in the wild with versatile real-world scenes.

\begin{table*}[htbp]
\centering
\footnotesize
\setlength\tabcolsep{0.7pt}
\scalebox{0.95}{
\begin{tabular}{
  >{\centering\arraybackslash}p{15pt}
  >{\centering\arraybackslash}p{60pt}
  >{\centering\arraybackslash}p{80pt}
  >{\centering\arraybackslash}p{65pt}
  >{\centering\arraybackslash}p{25pt}
  >{\centering\arraybackslash}p{60pt}
  >{\centering\arraybackslash}p{35pt}
  >{\centering\arraybackslash}p{35pt}
  >{\centering\arraybackslash}p{35pt}
  >{\centering\arraybackslash}p{35pt}
  >{\centering\arraybackslash}p{35pt}
  >{\centering\arraybackslash}p{35pt}
}
\toprule
\multicolumn{2}{c}{\multirow{2}{*}{Data type}} & \multirow{2}{*}{Datasets} & \multirow{2}{*}{Subjects} & \multirow{2}{*}{Scene} & \multirow{2}{*}{Frame}  & \multicolumn{6}{c}{Modalities} \\ 
\cline{7-12} 
\multicolumn{2}{c}{\rule{0pt}{8pt}}                           &       &       &  &  & RGB  & D/N & K2D & K3D & B.P. & WB.P. \\ 
\midrule

\rule{0pt}{8pt}\multirow{9}{*}{\rotatebox{270}{Real-world}} & \multirow{4}{*}{Monocular}       & COCO~\cite{coco}     & -  &  - & 104K &    \checkmark  &      &    \checkmark &     &            &          \\ 
\rule{0pt}{8pt}&         & MPI-INF-3DHP~\cite{mpi-inf3dhp}              & 8   & -& 1.4M  &  \checkmark    &      &    \checkmark  &   \checkmark   &            &          \\ 
\rule{0pt}{8pt}&         & MPII~\cite{mpii}    &  - & -& 24K&   \checkmark    &      &  \checkmark    &     &            &          \\ \cline{2-12} 
\rule{0pt}{8pt} & \multirow{3}{*}{Multi-view} & HuMMan       & 1000  & 1 & 60M&    \checkmark     &    \checkmark   & \checkmark     &\checkmark      &      \checkmark       &    \checkmark       \\ 

\rule{0pt}{8pt} &         & ZJU Mocap~\cite{neuralbody}       &6   & 1 & >1K&    \checkmark    &      &     \checkmark  &     \checkmark  &   \checkmark           &         \\ 
\rule{0pt}{8pt} &         & AIST++~\cite{aist++}     & 30  & 1 &10.1M &    \checkmark   &      &   \checkmark    &   \checkmark    &   \checkmark           &          \\ \cline{2-12} 
\rule{0pt}{8pt} & \multirow{2}{*}{MOCAP}      & 3DPW~\cite{3dpw}                  & 5 & <60  &  51K     &    \checkmark   &     &   \checkmark  &   \checkmark         &  \checkmark        \\
&         & Human3.6M~\cite{h36m}         &  11 & 1  &3.6M&    \checkmark   &    \checkmark    &   \checkmark    &  \checkmark    &          \checkmark    &          \\ \midrule
\multicolumn{2}{c}{\multirow{4}{*}{Synthetic}}                                & AGORA~\cite{agora}        &  >350 &  -& 18K&   \checkmark    &    \checkmark    &   \checkmark    &    \checkmark   &    \checkmark          &        \checkmark    \\
\multicolumn{2}{c}{\rule{0pt}{8pt}}   & Synbody~\cite{synbody}                   &  10K & 6 & 1.2M&   \checkmark     &    \checkmark    &    \checkmark   &   \checkmark    &        \checkmark      &      \checkmark      \\  
\multicolumn{2}{c}{\rule{0pt}{8pt}}   & BEDLAM~\cite{bedlam}                    & 217  & 103 & 280K &  \checkmark     &   \checkmark     &   \checkmark    &  \checkmark     &        \checkmark      &      \checkmark      \\ \midrule
\multicolumn{2}{c}{Generated}                                          & HumanWild                 &  $\infty$ & $\infty$   & 630K &   \checkmark   &    \checkmark    &   \checkmark    &    \checkmark   &        \checkmark      &    \checkmark        \\ \bottomrule

\end{tabular}
}
\caption{Comparison of different types of 3D HPS datasets.}
\label{tab:dataset_comparision}
\end{table*}

Our contributions can be summarised as follows. \textbf{1).} We curate a large-scale, human-centric dataset with comprehensive, high-quality annotations designed for controllable human image generation. \textbf{2).} Without pursuing fancy techniques, we propose a simple yet cost-efficient pipeline to synthesize realistic and diverse human images with well-aligned annotations, including paired SMPL-X parameters and human images. Beyond human pose and shape estimation, the dataset can empower a wide range of downstream perception tasks by rendering SMPL-X mesh into corresponding annotation format, \eg, human part segmentation. \textbf{3).} We verify the quality of the generated dataset on the 3D HPS task. 
Experimental results indicate that the proposed pipeline is compositional with CG-rendered data, enhancing performance across multiple challenging HPS benchmarks under consistent settings.

\section{Related Work}

\subsection{Human Pose and Shape Estimation Datasets} 
 \noindent\textbf{Real-world human pose data} plays a pivotal role in achieving precision and realism in 3D HPS tasks.
 High-quality 3D human data is typically captured using advanced motion capture devices like Inertial Measurement Units (IMUs)~\cite{3dpw, amass, humaneva} or Optical sensors~\cite{h36m}, designed to capture precise marker movements or joint rotations. 
Nevertheless, the utilization of these tools may present challenges attributable to factors such as financial expenses, intricacies in configuration, and spatial constraints.
 To facilitate these challenges, researchers have explored alternative methods to capture pseudo labels from diverse image types, including single-view images~\cite{smplify}, RGBD~\cite{prox}, and multi-view~\cite{humman}, eliminating the need for motion capture gear. SLOPER4D~\cite{Sloper4d} consolidates data from IMU sensors, LiDAR, and RGB information to construct a large-scale urban human-scene dataset. Such methods often leverage perception models to derive 2D cues from images, which are further optimized by a 3D joint re-projecting loss.

\myparagraph{Synthetic human pose datasets,} developed with computer graphics techniques, have been used for many years. 
SURREAL~\cite{surreal} applies human skin and cloth textures to bare SMPL meshes, which lack realistic details. AGORA~\cite{agora} uses high-quality static human scans for image rendering, but this routine also suffers from a high workload of scanning and rigging.
However, rendering realistic, manipulable synthetic human datasets involves many challenges, including the need for diverse virtual properties for realistic data. 
BEDLAM~\cite{bedlam} and Synbody~\cite{synbody} augment SMPL-X meshes~\cite{smplx} with diverse hair models and skin textures, facilitating the simulation of physically realistic clothes and hair dynamics.
These processes can be resource-intensive. Furthermore, the use of rendering engines demands many professional skills. Besides, the rendering process can be computationally expensive and time-consuming.

\myparagraph{Controllable human image generation} has gained great traction with the advancement of Stable Diffusion~\cite{stablediffusion, controlnet}. Text2Human~\cite{text2human} uses a diffusion-based transformer sampler in response to text prompts and predicts indices from a hierarchical texture-aware codebook to conditionally generate realistic human images.
HumanSD~\cite{humansd} introduces a skeleton-guided diffusion model with a novel heatmap loss for pose-conditioned human image generation. HyperHuman~\cite{hyperhuman} proposes to jointly denoise surface normal and depth along with the synthesized RGB image conditioned with text prompt and pose skeleton.

\myparagraph{Generative Models for Perception Tasks} is a rapidly growing field that aims to enhance the performance of perception models by utilizing the vast prior knowledge of generative models.
Several pioneering works have proposed generating perception datasets with diffusion models. 
For instance, \cite{synclass} and \cite{bigdatamyth} verified the effectiveness of datasets synthesized by generative models in fundamental perception tasks, \ie, image classification, and object detection.
StableRep~\cite{stablerep} argues that training modern self-supervised methods on synthetic images from Stable Diffusion Models can yield impressive results. The learned representations often surpass those learned from real images of the same sample size.
DatasetDM~\cite{datasetdm} leverages the prior of the Stable Diffusion model by training customized perception decoders upon the output of the UNet~\cite{unet} and then employs it to generate pseudo labels for semantic segmentation, depth estimation, and 2D human pose estimation.

\section{Method}\label{sec:method}

\begin{figure*}[t]
\begin{center}
   \includegraphics[width=0.96\linewidth]{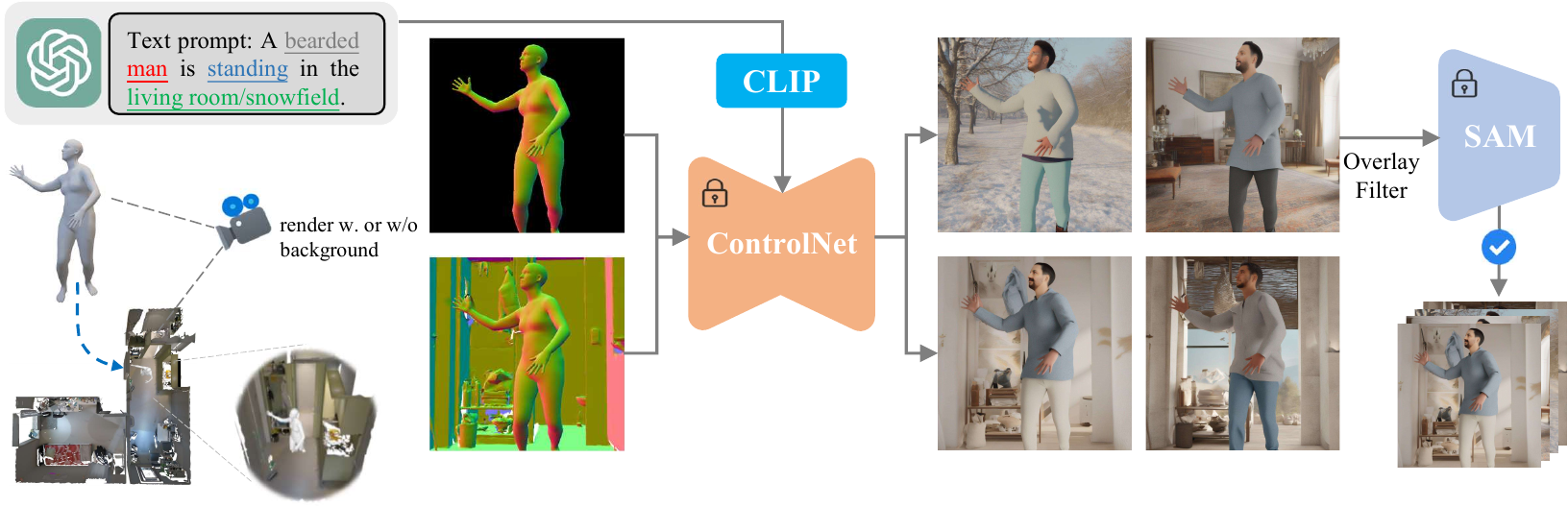}
\end{center}
\vspace{-0.0cm}
   \caption{
   The overall pipeline of the proposed controllable data generation. Our ControlNet could be conditioned on fine-grained keypoint maps, depth maps, surface normal maps, and structural text prompts. Our text prompt includes human appearance, pose, and indoor/outdoor scene types based on w. or w/o background normals. 
}
\vspace{-15pt}
\label{fig:data_pipeline}
\end{figure*}

 We present \Ours, a simple yet effective approach for creating versatile human body images and corresponding 3D parametric mesh annotations in a fully automated fashion, which can be used for many downstream human perception tasks, such as 2D/3D human pose and shape estimation (see~\cref{fig:data_pipeline}). The core idea of the proposed pipeline is creating large-scale image-mesh-caption pairs by incorporating 2D generative models, \eg, ControlNet~\cite{controlnet} and 3D human parametric models~\cite{smplx}. For the sake of completeness, we give a brief review of the controllable text-to-image (T2I), image-to-image (I2I) generative models, and the 3D human parametric model, SMPL-X \cite{smplx} in~\cref{method:prereq}. 
Then, we show how we build large-scale human-centric datasets in \cref{sec:human_dataset} for the controllable human generation task. Next, we illustrate how we generate the initial human image-annotation pairs in~\cref{method:init_gen}. Furthermore, we show our filtering strategy for the noisy labels to get high-quality training pairs in~\cref{method:label_denoise}.

\subsection{Prerequisites} \label{method:prereq}
\myparagraph{Stable Diffusion Models}    \cite{stablediffusion,sdxl} are text-to-image diffusion models capable of generating highly photo-realistic images given any text input. It consists of an autoencoder and a denoiser.
During training, U-Net model is designed to take a textual description and noise map $\epsilon$ as input, learning a forward noising process by gradually transforming an image latent variable, denoted as $\bar{y} \in \mathbb{R}^{h\times w \times c} $ into a noise map $\bar{y_t} \in \mathbb{R}^{h\times w \times c}$:
\begin{equation}
    y_t=\sqrt{\bar{\alpha}_t} y+\sqrt{1-\bar{\alpha}_t} \epsilon_t \quad \epsilon \sim \mathcal{N}(\mathbf{0}, \mathbf{I}),
    \label{eq:ddpm}
\end{equation}
where $t=[1, \cdots, T]$ is the timestep for controlling the noise level and ${\alpha}_t$ is the noise scheduler~\cite{ddpm}, $h$, $w$ and $c$ are the height, width, and channels of the latent variable. During inference, UNet takes a noise map as input and generates an approximation of the original latent variable by using a denoise scheduler~\cite{ddim}.

\myparagraph{SMPL-X}~\cite{smplx}, defined as $M(\boldsymbol{\beta},\boldsymbol{\theta}, \boldsymbol{\psi}): \mathbb{R}^{|\theta| \times|\beta| \times|\psi|} \rightarrow \mathbb{R}^{3 N}$,
is a 3D wholebody human parametric model, employing shape $\boldsymbol{\beta}\in\mathbb{R}^{200}$, expression $\boldsymbol{\psi}\in\mathbb{R}^{50}$, and pose $\boldsymbol{\theta}\in\mathbb{R}^{55\times3}$ to control the entire body mesh. 
The function of SMPL-X provides a differentiable skinning process that uses pose, shape, and expression parameters as inputs and delivers a triangulated mesh $V\in\mathbb{R}^{N\times3}$ with $N = 10475$ vertices. The reconstructed 3D joints $J\in\mathbb{R}^{144\times3}$ can be obtained using a forward kinematics process. We use the SMPL-X model to represent the human body.

\subsection{Human-centric Dataset for Human Generation}
\label{sec:human_dataset}

Large-scale datasets with high-quality samples, rich annotations, and diverse distributions are essential for controllable human image generation. To address this need, we have created a comprehensive human dataset with extensive annotations. Our dataset consists of two parts: CG-rendered human images and curated human images from existing sources, such as LAION-Human~\cite{laion5b} and other publicly available Internet datasets.

\noindent\textbf{Real-world Human.} For real-world human data collection, we filter images from LAION-2B-en~\cite{laion5b} and COYO-700~\cite{coyo-700m} using the YOLOS human detector~\cite{yolos}. Next, we apply state-of-the-art perception models—DepthAnything-V2~\cite{depthanythingv2}, DSINE~\cite{bae2024dsine}, and DWPose~\cite{dwpose}—to generate corresponding depth maps, surface normal maps, and keypoint heatmaps, respectively. The curated dataset covers a broad spectrum of scenes, including diverse backgrounds and partial human regions like clothing and limbs, providing a rich variety of real-world data for further use.

\begin{figure*}[t]
\begin{center}
   \includegraphics[width=0.96\linewidth]{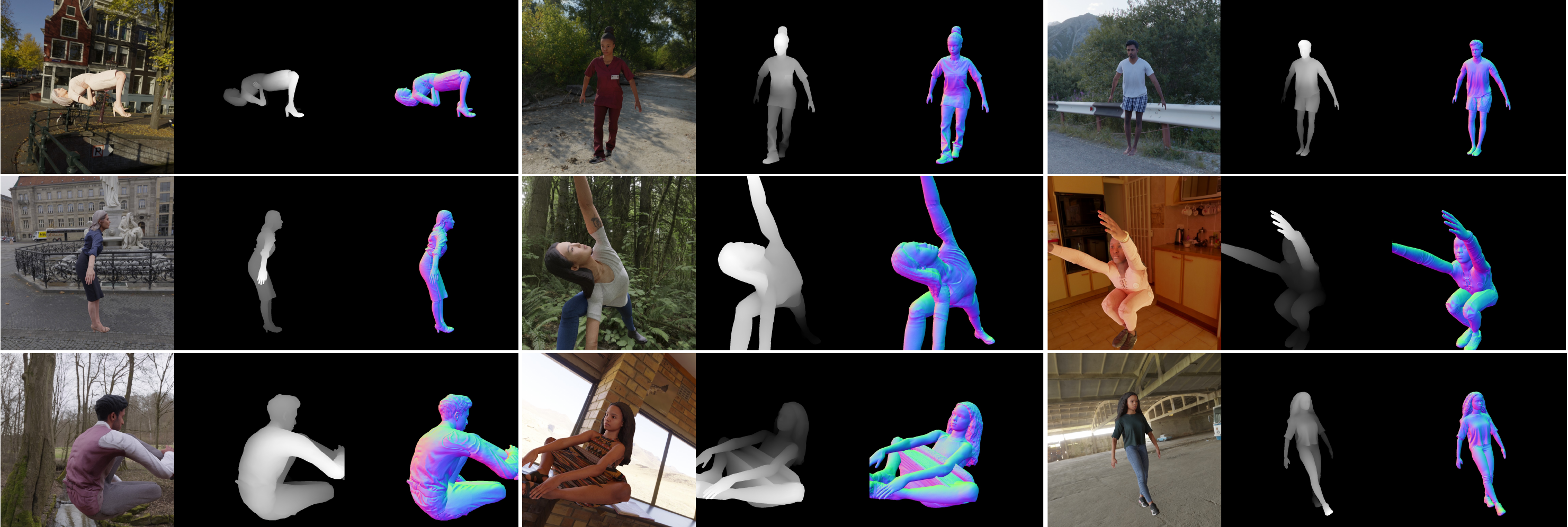}
\end{center}
\vspace{-0.0cm}
   \caption{
Visualization of our rendered human dataset with diverse human identities and in-the-wild scenes. Each data sample contains RGB image (left), a depth map (middle), and a surface normal map (right).
}
\vspace{-15pt}
\label{fig:renderpeople}
\end{figure*}

\noindent\textbf{CG-rendered Human.} We also introduce a synthetic dataset to provide accurate annotations for controllable human image generation. Specifically, we utilize 630 human models from RenderPeople~\cite{renderpeople} and 660 body pose sequences from AMASS~\cite{amass}. We use HDRi images with various lighting conditions as backgrounds. As shown in~\cref{fig:renderpeople}, the rendered dataset includes diverse human identities along with precise depth maps and surface normal maps without background, enhancing the dataset's utility for various tasks.

\begin{figure}[th]
    \centering
    \includegraphics[width=0.99\linewidth]{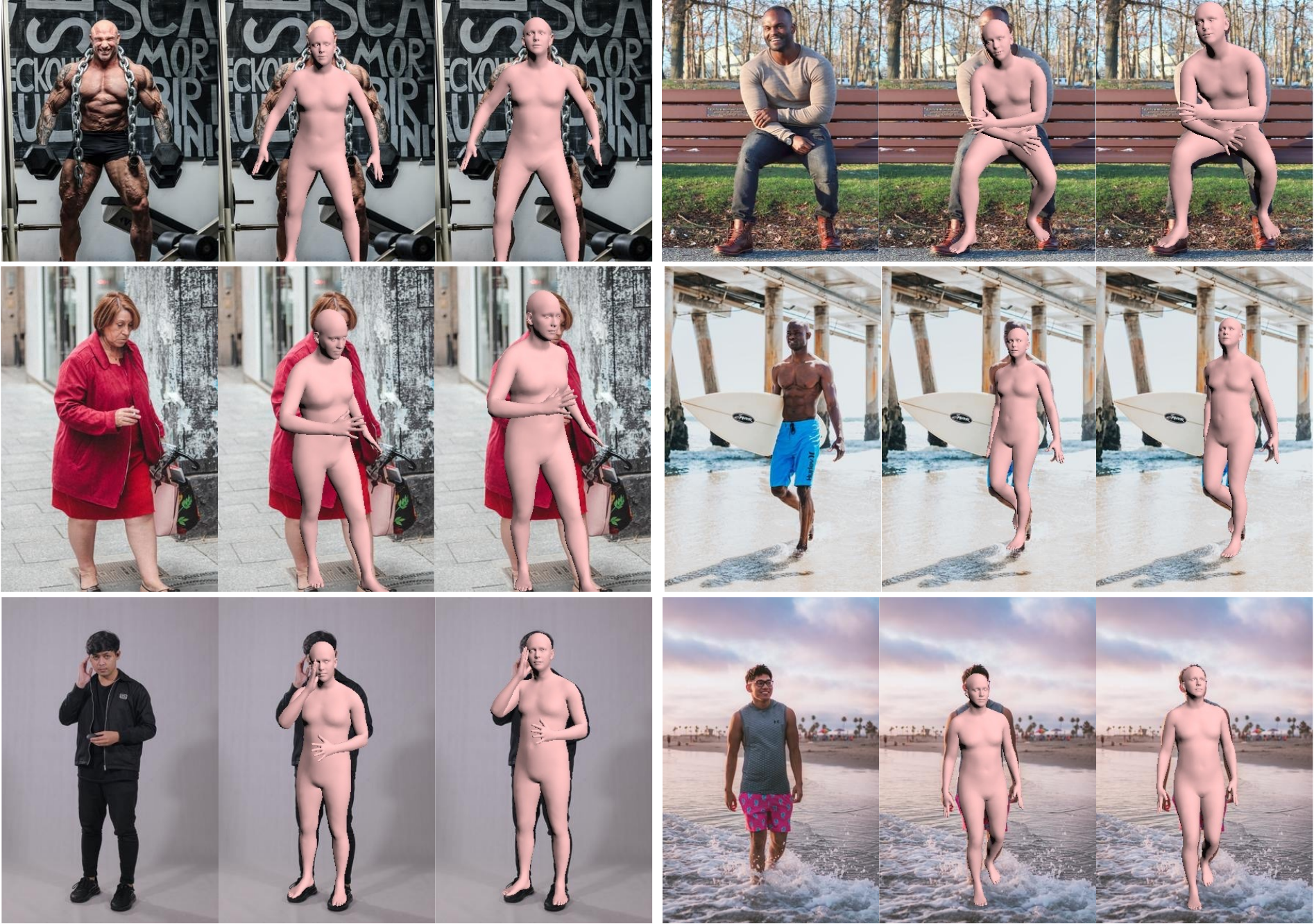}
    \caption{Visualization of HPS estimation. For each data sample, left is the original image, middle is trained on BEDLAM dataset, right is finetuned on our HumanWild dataset. }
\label{fig:hps_res_vis}
\end{figure}

\begin{figure*}[t]
\begin{center}
   \includegraphics[width=0.98\textwidth]{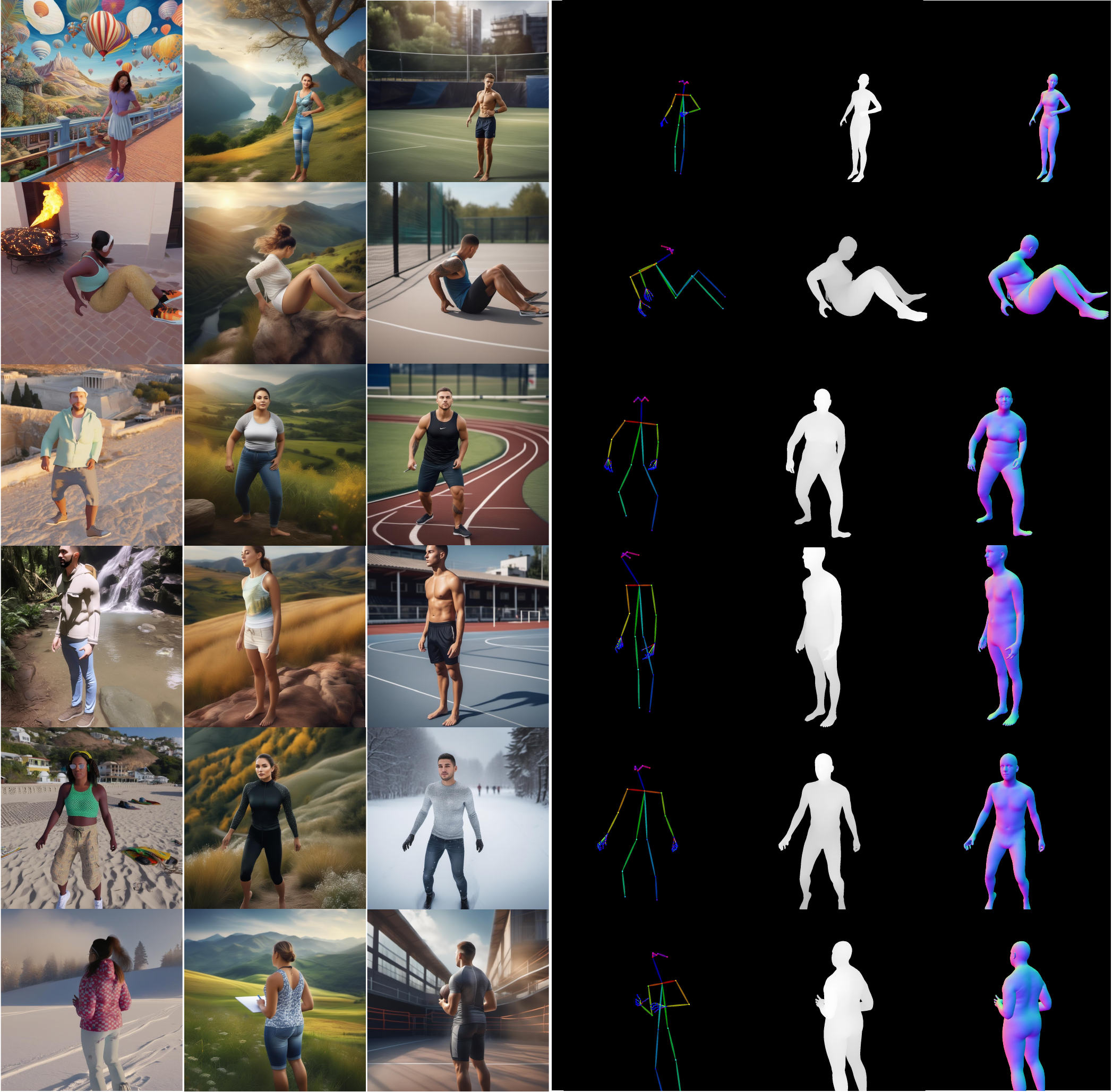}
\end{center}
\vspace{-0.0cm}
   \caption{
   \added[id=R2]{Visualization of human images generated by our multi-condition generation pipeline with diverse poses and scenes.} For each data sample, the first column is generated from the ControlNet before revision. \textbf{The second and third columns are generated from our revised model.} The fourth, fifth, and sixth columns are keypoint heatmap, depth map, and surface normal map rendered from the SMPL-X model.
   }
\vspace{-0.5pt}
\label{fig:single_human}
\end{figure*}

\noindent\textbf{Fine-tune Multi-condition ControlNet.} We then use the curated dataset for fine-tuning a multi-condition ControlNet~\cite{controlnet} based on SDXL~\cite{sdxl} for our customized controllable human image generation. The input conditions of ControlNet are text prompts, keypoint heatmaps, depth maps, and surface normal maps. As shown in~\cref{fig:single_human}, our fine-tuned ControlNet are generalizable to diverse poses and shapes generated by SMPL-X, while existing ControlNet variants trained on general datasets failed in many scenes.

\subsection{Initial Human Image and Annotation Generation} \label{method:init_gen}

\myparagraph{Camera simulation.} One drawback of vision-based motion capture systems is that they need to calibrate and synchronize the camera's intrinsic and extrinsic parameters during the capture. Thus, the collected human data are limited in terms of the scales and view diversity. On the contrary, our pipeline gets rid of the physical RGBD cameras and can simulate arbitrary human scales and body orientations.
Specifically, we randomly determine the orthographic scale $s$ of the human body, with  $s \in [0.45, 1.1] $. We also compute the horizontal shifts  $t_x $ and $ t_y $ within the ranges $[b_t/s, b_b/s]$ and $[b_l/s, b_r/s]$ respectively. Here, $b_t, b_b, b_l $, and $ b_r$  represent the empirical boundary values chosen based on whether a normal map is being generated for the half-body or the whole body.
Following \cite{hmr, zolly}, we determine the translation of the body as $transl = [t_x, t_y, f/s]$.
The focal length in normalized device coordinate (NDC) space, denoted as $f$, can be computed using the formula $f = 1 / \tan(FoV/2)$. Here, $FoV$ represents the Horizontal Field of View angle, which is randomly adjusted from 25 to 120 degrees by following \cite{bedlam, zolly}. {We also augment the human body's rotation by $[-180\degree, 180\degree]$ along $y$ axis and $[-30\degree, 30\degree]$ along $x, z$ axis.}

\myparagraph{Image condition generation.} To synthesize realistic human images with paired pose annotations, we leverage the multi-condition ControlNet fine-tuned in~\cref{sec:human_dataset}, as our image generator. Existing ControlNet variants usually take a 2D keypoint heatmap, depth map, or surface normal map as condition inputs. These inputs are typically detected from real-world images by pre-trained perception models. Different from the common usage, we construct the input of ControlNet by taking advantage of the 3D human parametric model, SMPL-X.
Specifically, we use existing large-scale human motion capture databases~\cite{amass} with diverse body poses and shapes in SMPL-X format to render the spatial input condition maps, \ie, keypoint heatmaps, depth maps, and surface normal maps. 
Thanks to the disentanglement of pose and shape parameters in the SMPL-X model, we can even recombine these parameters to generate a human mesh that doesn't exist in the databases.

Upon getting the simulated camera parameters aforementioned in~\cref{method:init_gen} and 3D human mesh from SMPL-X, we can render an existing 3D human mesh into the image plane by using perspective projection, as such, getting the corresponding source normal map, depth map, and 2D keypoint heatmap. 
Notably, the surface normal map is proven to be crucial to generating accurate body shapes and orientation, and the keypoint heatmap and depth map are useful for improving the alignment between the input body pose and the generated image.

\myparagraph{Human $\&$ scene positioning.}
We offer an optional flexible background selection feature, allowing users to generate scenes based on text prompts solely or in conjunction with background-free keypoint heatmaps, depth maps, and normal maps. To integrate specific backgrounds, we randomly select scene meshes from the ScanNet++ dataset~\cite{scannet++}. We first utilize Octree~\cite{octree} to partition the room mesh into discrete voxels. With vertex-level classification annotations, we efficiently identify ground plane normals and height to anchor the human mesh. We randomly allocate human mesh within unoccupied areas of the voxel space. Subsequently, we refine human mesh positioning by optimizing the translation using the Chamfer distance to preclude inter-mesh collisions. Finally, we simulate a random camera perspective. The intrinsic matrix is obtained with the previous rules, while the extrinsic matrix is crafted using a random azimuth angle. We sample the camera's height range [-1m, +1m] relative to the pelvis height, and look at a random point on the human torso, to achieve a realistic viewpoint. The visualization results can be seen in the appendix.

\myparagraph{Text prompt generation.} To create controllable and diverse in-the-wild background scenes, we can use text prompts to provide background information. Moreover, the human body's spatial maps are not fine-grained enough to determine the gender and appearance of a human. Thus, we incorporate a structured text prompt template to handle this issue. In particular, we designed a simple text template as ``A \{gender\} \{action\} \{environment\}". The gender of the person is determined by the SMPL-X annotations. The environment is generated by a large language model, \ie, ChatGPT~\cite{gpt3} and LLAMA~\cite{llama}. To create photo-realistic humans, we also feed negative text prompts, \eg, ``ugly, extra limbs, poorly drawn face, poorly drawn hands, poorly drawn feet'', to the model.

Finally, we have all of the input conditions of the multi-condition ControlNet. We apply a total of 40 inference steps for each sample. The generated images are up to $1024 \times 1024$ resolution. The generated images and the input conditions (SMPL-X parameters) are regarded as the initial data pairs.

\subsection{Label Denoising} \label{method:label_denoise}
\begin{figure*}[ht!]
    \centering
    \includegraphics[width=0.99\linewidth]{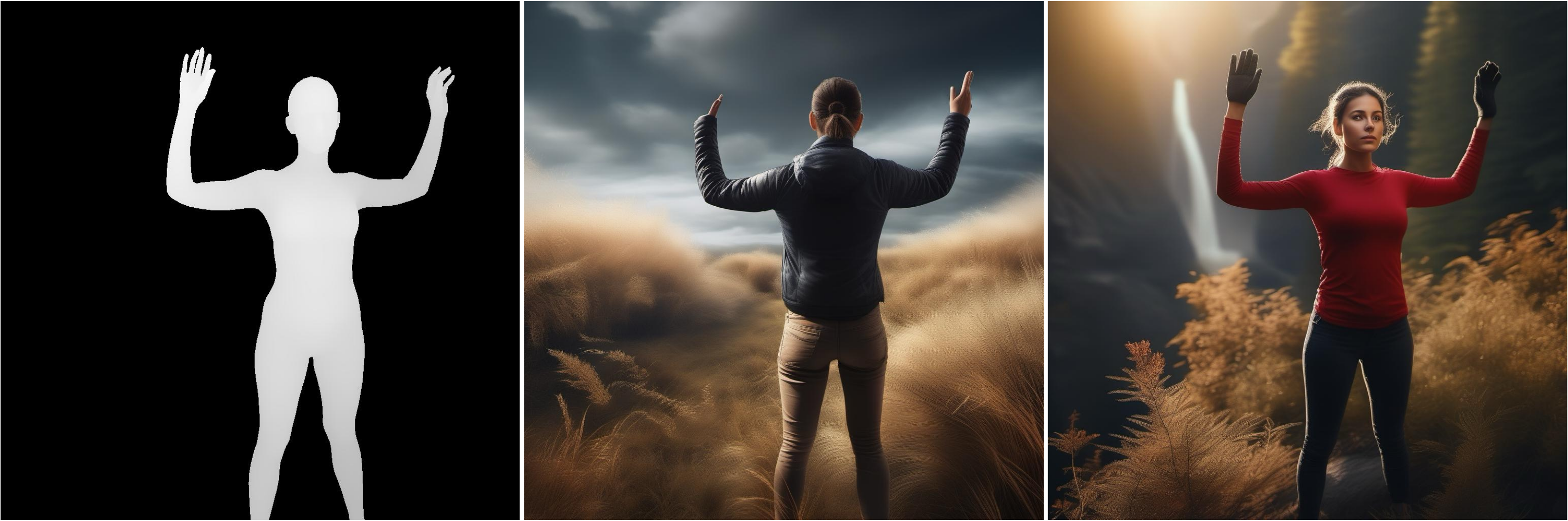}
    \caption{Analysis of Mirror Image Generation from Depth Maps: This visualization demonstrates the correction of a common failure mode. The leftmost image is the guiding depth map. The middle image is an incorrect output where the model mistakenly generates the person's back. The rightmost image shows the desired outcome, a correctly rendered frontal view. }
\label{fig:mirror_pair}
\end{figure*}

The generated images are not always well-aligned with the input conditions. For example, an incorrect case is that the generated human and the input conditions form a mirror pair, as shown in~\cref{fig:mirror_pair}. To resolve this problem, we employ an off-the-shelf foundation segmentation model, \ie, SAM~\cite{sam}, to filter negative samples from the final dataset. Specifically, we first calculate the ground-truth mesh segmentation mask by rendering the 3D mesh into the image plane. Then, we sample a random point coordinate from the ground-truth segmentation mask and feed the generated image and this point coordinate into the SAM model, thereby we can get the predicted human mask for the generated image. Finally, we compute the intersection-over-union (IoU) metric between the ground-truth mask and the prediction mask. We filter data samples with IoU lower than $0.8$.

To further improve the alignment between the generated human images and SMPL-X pose parameters, we train a sparse 2D surface estimator by using RTMPose~\cite{rtmpose} architecture to estimate a total of 128 sparse 2D vertices sampled from dense SMPL-X vertices~\cite{virtual_marker}. The input resolution of the detector is $512\times384$. As our generated images have accurate camera intrinsics, we can project the ground-truth 3D SMPL-X vertices into 2D vertices using perspective projection. Then, we can optimize the original SMPL-X pose parameters with SMPLify~\cite{smplify}. Note that we only optimize the body pose parameters for 20 iterations and make the shape parameters fixed.

\section{Experiments}

\subsection{Human-centered Datasets for training ControlNet}
\added[id=R2]{
In this section, we list our datasets for fine-tuning the multi-condition ControlNet in terms of the number of subjects, the number of scenes, and the details of diversity. Overall, we use three datasets for fine-tuning an existing variant. The RenderedHuman dataset contains 202,667 human images, distinguishing itself by providing an unparalleled wealth of precise annotations, including texts, keypoints, depth, and surface maps. The PexHuman is collected with more than 120 daily human activities, including 100,880 images. We also filter the Laion5B dataset by a human detector, YOLOS~\cite{yolos}, where only images with fewer than 3 humans are kept. We generate keypoints, depth, and surface maps for PexelHuman and LaionHuman with DWPose~\cite{dwpose}, DepthAnythingV2~\cite{depth_controlnet}, and DSINE~\cite{bae2024dsine}, separately. To get the text description of the human images, we leverage a multi-modality vision-language model, CogVLM~\cite{wang2023cogvlm}, to generate detailed image captions. 
}
\begin{table}[!t]
\centering
\footnotesize
\resizebox{0.45\textwidth}{!}{
\begin{tabular}{l|c|c|c|c}
\toprule
Dataset  & Data Type & \# Samples & \# Scenes & \# Identies \\ 
\midrule
LaionHuman & real & 233,728 & > 200K & > 200K \\
PexelHuman & real & 100,880 & > 100K & > 100K \\
RenderHuman & Synthetic & 202,667 & 660 & 630 \\
\bottomrule
\end{tabular}\textbf{}
}
\caption{{An overview of all datasets for fine-tuning multi-condition ControlNet.}}

\label{tab:train_data}
\end{table}

\subsection{Implement Details and Evaluation Metrics}
To illustrate the effectiveness and efficiency of our proposed \Ours, we report the Frechet Inception Distance (FID~\cite{heusel2017gans}) and the Kernel Inception Distance (KID~\cite{binkowski2018demystifying}), which are widely used to measure the quality of the synthesized images.
We measure the alignment accuracy between the SMPLX-rendered mask and the generated images with mean intersection over union (mIoU). 
We use standard metrics to evaluate body pose and shape accuracy. PVE and MPJPE represent the average error in vertices and joints positions, respectively, after aligning the pelvis. PA-MPJPE further aligns the rotation and scale before computing distance. PVE-T-SC is a per-vertex error in a neutral pose (T-pose) after scale correction. For a fair comparison with the counterpart pure CG pipeline, we sample SMPL-X pose and shape parameters from the AMASS~\cite{amass} dataset and render them into surface normal maps with random camera parameters illustrated in~\cref{sec:method}. We re-implemented a regression-based HPS method, CLIFF~\cite{cliff}, for evaluating the effectiveness of the synthesized dataset on a variety of evaluation benchmarks, 3DPW~\cite{3dpw}, AGORA~\cite{agora}, EgoBody~\cite{egobody}, RICH~\cite{rich}, and SSP-3D~\cite{ssp3d} (for shape evaluation).

Unless specified, all HPS models are trained on the CLIFF~\cite{cliff} model with an HRNet-48~\cite{hrnet} backbone. Following BEDLAM~\cite{bedlam}, the supervision losses are a combination of MSE loss on model parameters, projected keypoints, 3D joints, and an L1 loss on 3D vertices.

\subsection{Ablation on ControlNet Input Conditions}

In the initial stages of our research, we employed various input conditions, including 2D keypoint heatmaps, depth images, and normal images, to generate data samples. However, a notable challenge emerged with regard to the ambiguity inherent in both the keypoint and depth conditions. Specifically, the diffusion model struggled to accurately discern the front and back of individuals depicted in these data modalities and had difficulty in aligning generated hands and faces with ground-truths (See Sup. Mat. for some failure cases),
yet we found that surface normal conditioning can greatly reduce this kind of ambiguity. Thus, we choos a multi-condition variant of the ControlNet, which can optionally treat keypoint, depth, and surface normal maps as input. 
\cref{table:controlnet_eval} compares our trained ControlNet with two publicly available keypoint-based and depth-based ControlNets~\cite{depth_controlnet, pose_controlnet} in terms of generated image quality and image/ground-truth alignment accuracy. In order to ensure equitable comparison, all ControlNets employ aligned sets of 1,000 input conditions derived from identical SMPL-X parameters and camera specifications.
The results demonstrate that our ControlNet consistently outperforms the keypoint-based and depth-based ControlNet.

\begin{table}[htbp]
    \centering
    \scalebox{0.99}{
    \begin{tabular}{lccc}
    \toprule
     ControlNet & \multicolumn{2}{c}{Image Quality} &  \multicolumn{1}{c}{Alignment Accuracy}  \\
    \cmidrule(r){2-3} \cmidrule(r){4-4}
    Condition & FID $\downarrow$ & $\text{KID} \downarrow$  & mIoU $\uparrow$\\
    \midrule
    Keypoint~\cite{pose_controlnet} & 29.6 & 2.92  & 41.3 \\
    Depth~\cite{depth_controlnet} & 28.1  & 2.87  & 49.2 \\
    Normal & {24.5} & {2.83} & {56.8} \\
    All conditions & \textbf{22.5} & \textbf{2.74} & \textbf{59.2} \\
    \bottomrule
    \end{tabular}
    }
    \caption{Ablation of different input conditions of ControlNet.}
    \label{table:controlnet_eval}
\end{table}

\begin{figure*}
    \centering
    \includegraphics[width=0.9\linewidth]{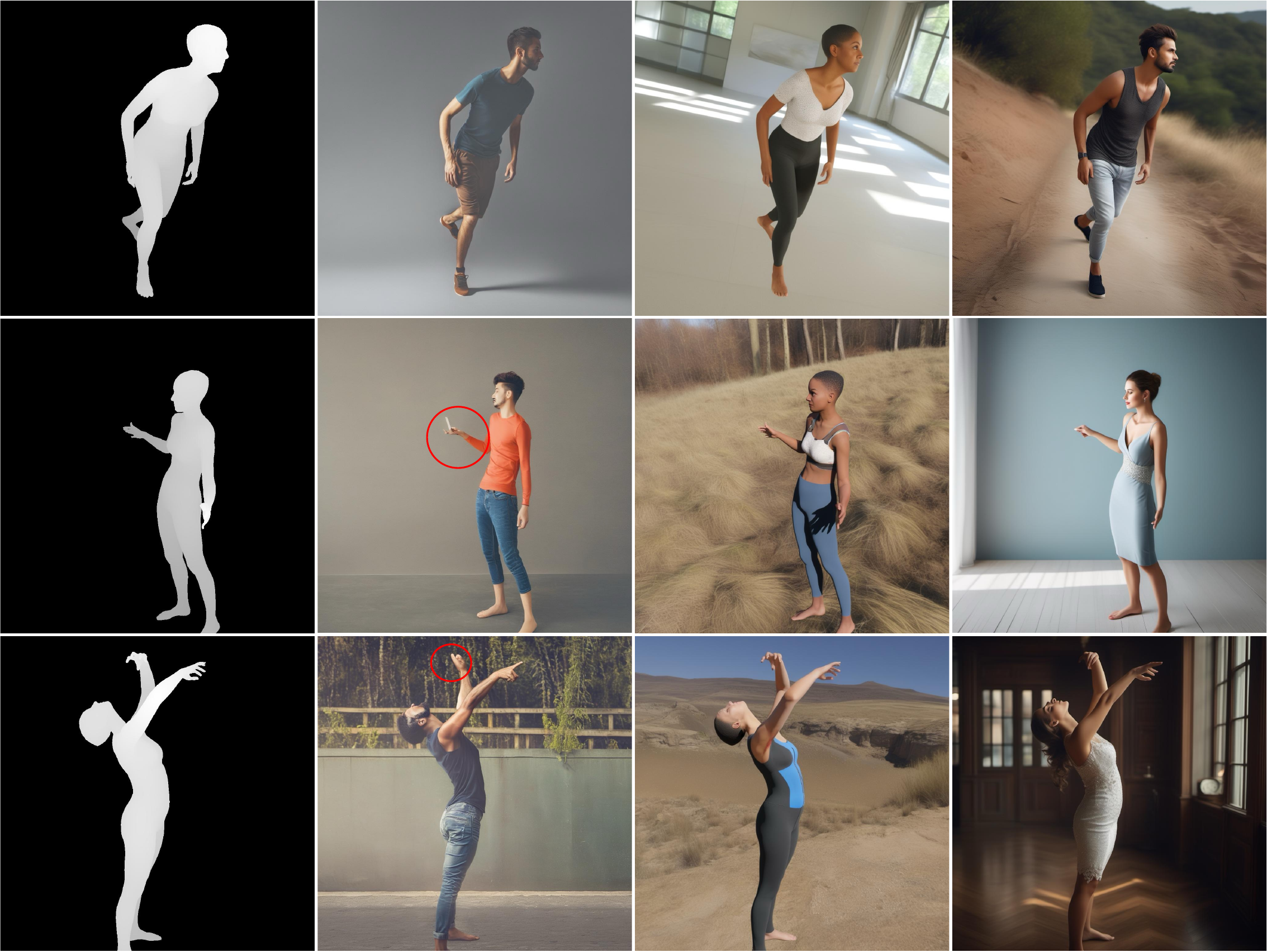}
    \caption{\added[id=R1,id=R2]{Visualization of a general depth-conditioned ControlNet, and our finetuned multi-condition ControlNet. The first column is the input depth condition. The second column is the prediction from the general depth-conditioned ControlNet. The third column is our v1 model. The fourth column is our current v2 model.}}
    \label{fig:compare_controlnet}
\end{figure*}

We also visualize the generated images by a general depth-conditioned ControlNet, and our finetuned multi-condition ControlNet in~\cref{fig:compare_controlnet}. As we can see, the general depth-conditioned ControlNet trained on general datasets fails to generate accurate fingers, poses, and limited background scenes, while our fine-tuned multi-condition ControlNet can generate diverse photo-realistic humans tightly aligned with the input condition.

\subsection{Verification of the Necessity of In-the-wild 3D %
Data}
Previous studies, HMR-Benchmark~\cite{hmr-benchmark} and BEDLAM~\cite{bedlam} point out that different backbone initialization strategies profoundly impact an HPS model's convergence speed and final accuracy. HMR-benchmarks~\cite{hmr-benchmark} conduct a systematic analysis by initializing the same backbone pre-trained with different sources, \ie, ImageNet classification weights, 2D pose estimation weights from MPII, and COCO. As shown in the first three columns of \cref{tab:pretrain_ablation} show the results are trained on real datasets. The conclusion is that COCO keypoint pretraining helps HPS models trained on real datasets achieve the best performance.

To verify whether the conclusion still holds for CG-rendered synthetic datasets, columns 4 to 6 of the \cref{tab:pretrain_ablation} elucidate the impact of backbone pretraining on BEDLAM~\cite{bedlam} and AGORA~\cite{agora}. A consistent conclusion is drawn that COCO keypoint pretraining also has a great positive impact on CG-rendered datasets. 

Based on the above observation, we posit that neither indoor-mocap datasets nor CG-rendered datasets offer a sufficiently varied array of photo-realistic, in-the-wild scenes. Consequently, the HPS network, which is not initialized with large-scale in-the-wild data pretraining, struggles to perform well on unseen scenes and has limited generalization capability. 
As our generated data contains diverse in-the-wild scenes, we add the data generated by our pipeline for training HPS models. As we can see in the final column of ~\cref{tab:pretrain_ablation}, the inclusion of in-the-wild synthesized data generated by \Ours can further enhance the HPS performance of the CG-rendered datasets.
Hence, data synthesized by diffusion models can effectively complement CG-rendered data by featuring diverse human identities and real-world scenes.

\begin{table}[htbp]
    \vspace{0pt}
    \centering
    \begin{minipage}[t]{\linewidth} %
        \centering
        \scalebox{0.96}{
\begin{tabular}{L{65pt}M{55pt}M{50pt}M{40pt}}
    \toprule
    Training Data
     & 
    \multicolumn{1}{l}{\footnotesize PA-MPJPE$\downarrow$} &
    \multicolumn{1}{l}{\footnotesize MPJPE$\downarrow$} &
    \multicolumn{1}{l}{\footnotesize PVE$\downarrow$} \\
    \midrule
    PDHuman &102.4 &159.1 &168.0 \\
    \Ours & \textbf{90.0} & \textbf{151.8} & \textbf{160.9} \\
\bottomrule
\end{tabular}
}
        \caption{HPS results on SPEC-MTP.}
        \label{tab:spec_mtp}
    \vspace{15pt} %
    \end{minipage}
    \begin{minipage}[t]{\linewidth} %
        \centering
        \scalebox{0.96}{\begin{tabular}{
 L{65pt}cccc}
\toprule[1pt]
Training Data       & MPJPE~$\downarrow$ & PA-MPJPE~$\downarrow$& PVE~$\downarrow$& PA-PVE~$\downarrow$ \\
\midrule
 B+A+P   & 143.1 & 94.9 & 163.9 & 95.7 \\
 B+A+P+H  & \textbf{134.6}  & \textbf{89.1} & \textbf{154.1} & \textbf{88.9} \\
\bottomrule[1pt] 
\end{tabular}
}
        \caption{HPS results on MOYO.}
        \label{tab:moyo}
    \end{minipage}
    \vspace{-0pt}
\end{table}

\begin{table*}[t]
	\centering
 
   \scalebox{1.0}{
 \begin{tabular}{l|l|c|ccc|c}

	\toprule
	\multicolumn{1}{c|}{Method} & \multicolumn{1}{c|}{Dataset} & \multicolumn{1}{c|}{Pretrain}  & \multicolumn{1}{c}{PA-MPJPE$\downarrow$} & \multicolumn{1}{c}{MPJPE$\downarrow$} & \multicolumn{1}{c|}{PVE$\downarrow$} & \multicolumn{1}{c}{PVE-T-SC$\downarrow$} \\
        \midrule
        PARE~\cite{pare,hmr-benchmark} & R & ImageNet  & 54.8 & N/A  & N/A  &  N/A \\
        PARE~\cite{pare,hmr-benchmark} & R & MPII  & 51.5 & N/A  & N/A &  N/A \\
        PARE~\cite{pare,hmr-benchmark} & R & COCO  & 49.5 & N/A  & N/A  &  N/A \\
	\midrule
	CLIFF~\cite{cliff,bedlam} & B+A & scratch  & 61.7 & 96.5 & 115.0 & N/A \\
	CLIFF~\cite{cliff,bedlam} &  B+A & ImageNet  & 51.8 & 82.1 & 96.9 & N/A \\
	CLIFF~\cite{cliff,bedlam} &  B+A & COCO  & 47.4 & 73.0 & 86.6 & 14.2 \\
        \midrule
        CLIFF~\cite{cliff,bedlam} &  B+A+\textbf{H} & COCO   & \textbf{44.9} &  \textbf{70.2} & \textbf{82.7} & \textbf{13.9} \\
\bottomrule
\end{tabular}
}
\caption{Ablation experiments on 3D pose and shape estimation. `R' denotes mixed real-world datasets, `H' denotes \Ours, `B' denotes BEDLAM and `A' denotes AGORA. PA-MPJPE, MPJPE and PVE are evaluated on 3DPW. PVE-T-SC is evaluated on SSP-3D.}
\label{tab:pretrain_ablation}
\end{table*}

\subsection{Data Scale-up Ablation Study}
\begin{table*}[t]
  \centering
  \scalebox{0.9}{
  \begin{tabular}{>{\centering\arraybackslash}p{45pt}>{\centering\arraybackslash}p{45pt}>{\centering\arraybackslash}p{45pt}>{\centering\arraybackslash}p{45pt}|ccccc|>{\centering\arraybackslash}p{40pt}}
    \toprule
    \#Crops & \#Inst. & Model & \#Param. &
    AGORA~\cite{agora} &
    EgoBody~\cite{egobody} &
    RICH~\cite{osx} & 
    3DPW~\cite{3dpw} &
    H36M~\cite{h36m} & 
    MPE \\
    
    \midrule
    25\% & 0.19M & ViT-S & 23M & 125.0 & 114.2 & 116.5 & 118.2 & 102.3 & 115.2 \\
    50\% & 0.39M & ViT-S & 23M & 116.2 & 103.6 & 104.1 & 110.4 & 95.6  & 106.0 \\
    \midrule
    100\% & 0.63M & ViT-S & 23M & 106.8 & 97.3 & 107.7 & 106.5 & 87.1 & 101.1 \\
    100\% & 0.63M & ViT-B  & 88M & 103.6 & 94.2 & 105.1 & 103.5 & 82.5 & 97.8 \\
    100\% & 0.63M & ViT-L & 305M & 97.5 & 85.6 & 97.2 & 97.4 & 75.7 & 90.7 \\
    100\% & 0.63M & ViT-H & 633M & 90.2 & 82.3 & 92.1 & 90.3 & 72.4 & 85.5 \\
    \bottomrule
    \end{tabular}}
  \caption{\textbf{Scale-up ablation.} We study the scaling law of the amount of data and the model sizes. The metrics are MPJPE for Human36M~\cite{h36m} and PVE for other evaluation benchmarks. 
  Foundation models are named ``ViT-M", where M indicates the size of ViT backbone (S, B, L, H).
  MPE: mean primary error cross 5 evaluation benchmarks. %
  AGORA uses the validation set, and EgoBody uses the EgoSet. 
  }
    \vspace{0cm}
  \label{tab:scale_up}
\end{table*}

We perform experiments to show the effectiveness of the generated datasets by studying the scaling law of the amount of data and the model sizes in~\cref{tab:scale_up}. Notably, our training set is independent of the evaluation benchmarks in~\cref{tab:scale_up}. We show results with various ViT backbones and percentages of the proposed dataset. It is observed that 1). Scale-up synthetic training data generated by \Ours leads to better performance. 2). A larger backbone benifits more from a larger synthetic training dataset.

\subsection{Comparing with Other Real/Synthetic Datasets}
In Table \ref{tab:3dpw-rich-compare}, we analyze the performance of different training datasets using mixed strategies on CLIFF with HRNet-W48 backbone. Key findings are: 
1). Pure CG-rendered dataset outperforms real-world data (motion-capture, pseudo-labeled) on 3DPW and RICH datasets, suggesting the effectiveness of the synthetic data.
2). Solely training on HumanWild yields slightly lower performance than BEDLAM; we conjecture that there still exists some levels of label noise, especially in hand and face poses.
3). Combining HumanWild with CG-rendered datasets (BEDLAM, AGORA) and marker-based dataset (3DPW) enhances performance, which demonstrates the flexibility of \Ours, \ie, can be integrated with existing datasets to boost the estimation performance.

These results provide additional validation for the efficacy of the generated HumanWild dataset. Specifically, they demonstrate that incorporating HumanWild data enables the HPS regressor to encounter a more diverse range of scenarios, thereby enhancing its ability to generalize to in-the-wild scenes.

\begin{table*}[htbp]
\centering

\scalebox{1.2}{
\begin{tabular}{ r |c|ccc|ccc}
    \toprule
    \multirow{2}{*}{Methods} & 
    \multirow{2}{*}{Training Data} & 
    \multicolumn{3}{c|}{3DPW (14) }&         
    \multicolumn{3}{c}{RICH (24)} \\
    & & \multicolumn{1}{l}{\footnotesize PA-MPJPE$\downarrow$} & \multicolumn{1}{l}{\footnotesize MPJPE$\downarrow$} & 
    \multicolumn{1}{l|}{\footnotesize PVE$\downarrow$} & 
    \multicolumn{1}{l}{\footnotesize PA-MPJPE$\downarrow$} &
    \multicolumn{1}{l}{\footnotesize MPJPE$\downarrow$} &
    \multicolumn{1}{l}{\footnotesize PVE$\downarrow$} \\
    \midrule
    HMR        \cite{hmr} & Real & 76.7 & 130 & N/A & 90.0 & 158.3 & 186.0 \\
    SPIN        \cite{spin} & Real & 59.2	& 96.9 &116.4 &69.7& 122.9 & 144.2\\
    SPEC        \cite{spec} & Real	&53.2 &	96.5 &	118.5 & 72.5 &127.5 & 146.5 \\
    PARE        \cite{pare} & Real &50.9 & 82.0 & 97.9 & 64.9 & 104.0 & 119.7 \\
    HybrIK        \cite{hybrik} & Real & 48.8 &	80 & 94.5 & 56.4 & 96.8 & 110.4 \\
        CLIFF \cite{cliff,bedlam} & Real & {46.4} &73.9 & 87.6 & 55.7 & 90.0 & 102.0 \\
       \midrule
    CLIFF \cite{cliff,bedlam}  & B & 50.5 & 76.1 & 90.6 & - & - & - \\
    CLIFF \cite{cliff,bedlam}  & \textbf{H} & 51.4 & 83.9 & 96.4 & - & - & - \\
       \midrule
    CLIFF \cite{cliff,bedlam}  & B+A & 46.6 & {72.0} & {85.0} & {51.2} & {84.5} & {96.6}\\
    CLIFF \cite{cliff,bedlam}  & \textbf{H}+A & 47.5 & {74.1} & {87.8} & {51.9} & {85.3} & {97.4}\\
        CLIFF\cite{cliff,bedlam} & \textbf{H}+B+A &\textbf{44.9}  & \textbf{70.4}  & \textbf{82.0} & \textbf{51.0} & \textbf{84.4} & \textbf{96.1} \\
        \midrule
       CLIFF\cite{cliff,bedlam} & B+A+P& {43.0} & {66.9} & {78.5} & {50.2} & {84.4} & {95.6}\\
        CLIFF\cite{cliff,bedlam}  & \textbf{H}+B+A+P  &\textbf{41.3} & \textbf{65.2} & \textbf{76.7} & \textbf{48.4} & \textbf{79.7} & \textbf{91.1}\\		
        CLIFF\cite{cliff,bedlam}  & \textbf{H}+B+A+P+R  &{45.5} & {71.3} & {84.3} & {52.6} & {86.4} & {98.3} \\		 
        \bottomrule
\end{tabular}
\vspace{-0.1cm}

}
\caption{Reconstruction error on 3DPW and RICH. `R' denotes mixed real-world datasets, `H' denotes \Ours, `B' denotes BEDLAM~\cite{bedlam}, and `A' denotes AGORA~\cite{agora}. `P' denotes 3DPW~\cite{3dpw}.}
\label{tab:3dpw-rich-compare}
\end{table*}

\subsection{Result on Challenging Benchmarks}
To verify the effectiveness of \Ours on different scenerios, we report the detailed performance on three challenging benchmarks. 
Firstly, we evaluate the capability of \Ours in handling perspective distortion human pose and shape estimation, by performing fair ablation using Zolly~\cite{zolly} with ResNet50~\cite{resnet} backbone. The results in~\cref{tab:spec_mtp} show that \Ours dataset achieves higher generalization performance on SPEC-MTP~\cite{spec} when compared with PDhuman~\cite{zolly}, which is a CG-rendered perspective distortion dataset. Secondly, MOYO~\cite{moyo} encompasses challenging yoga poses not present in BEDLAM and \Ours datasets, providing a unique testbed for evaluating the model's generalization ability on hard poses. The results in~\cref{tab:moyo} demonstrate that \Ours, complements existing datasets, resulting in improved generalization performance on the MOYO test set.

We show AGORA test set results in~\cref{tab:agora_test}, our analysis reveals that while Humanwild demonstrates superior performance over BEDLAM on body metrics, it exhibits relatively poorer performance on hand and face metrics, particularly evident in comparison to the AGORA test set. We hypothesize that the alignment of hand and facial annotations with generated images remains noisy, primarily due to inherent limitations in current diffusion models' ability to accurately generate hand poses, particularly within small resolutions.
\begin{table*}
  \centering
  \vspace{-2mm}
  \resizebox{\textwidth}{!}{
  \begin{tabular}{lccccccccccccccc}
    \toprule
    & 
    \multicolumn{2}{c}{NMVE$\downarrow$ (\emph{mm})} &
    \multicolumn{2}{c}{NMJE$\downarrow$ (\emph{mm})} &
    \multicolumn{5}{c}{MVE$\downarrow$ (\emph{mm})} &
    \multicolumn{5}{c}{MPJPE$\downarrow$ (\emph{mm})} \\
    
    \cmidrule(lr){2-3} \cmidrule(lr){4-5} \cmidrule(lr){6-10} \cmidrule(lr){11-15}  
    Method &
    All & Body &
    All & Body &
    All & Body & Face & LHand & RHand &
    All & Body & Face & LHand & RHhand 
    \\
    \midrule
    BEDLAM~\cite{bedlam}& 
    179.5 & 132.2 & 177.5 & 131.4 & 131.0 & 96.5 & \underline{25.8} & \underline{38.8} & \underline{39.0} & 129.6 & 95.9 & \underline{27.8} & \underline{36.6} & \underline{36.7} \\
    
    Hand4Whole~\cite{hand4whole}$\dagger$& 
    144.1 & 96.0 & 141.1 & 92.7 & 135.5 & 90.2 & 41.6 & 46.3 & 48.1 & 132.6 & 87.1 & 46.1 & 44.3 & 46.2 \\
    
    BEDLAM~\cite{bedlam}$\dagger$ & 
    142.2 & 102.1 & 
    141.0 & 101.8 & 
    \textbf{\underline{103.8}} & 74.5 & \textbf{23.1} & \textbf{31.7} & 
    \textbf{33.2} & \textbf{\underline{102.9}} & 74.3 & \textbf{24.7} & \textbf{29.9} & \textbf{31.3} \\
    
    PyMaF-X~\cite{pymafx}$\dagger$ &
    141.2 & 94.4 & 
    140.0 & 93.5 & 
    125.7 & 84.0 & 35.0 & 44.6 & 
    45.6 & 124.6 & 83.2 & 37.9 & 42.5 & 43.7 \\
    
    OSX~\cite{osx}$\ast$&
    130.6 & 85.3 & 
    127.6 & 83.3 & 
    122.8 & 80.2 & 36.2 & 45.4 & 
    46.1 & 119.9 & 78.3 & 37.9 & 43.0 & 43.9 \\

    \Ours$\dagger$& 
    \textbf{120.5} & \textbf{73.7} & 
    \textbf{115.7} & \textbf{72.3} & 
    {112.1} & \textbf{68.5} & {37.0} & {46.7} &
    {47.0} & {107.6} & \textbf{67.2} & {38.5} & {41.2} & 41.4 \\
    
    \bottomrule

  \end{tabular}
  }
\caption{\textbf{AGORA test set.} $\dagger$ denotes the methods that are finetuned on the AGORA training set. $\ast$ denotes the methods that are trained on AGORA training set only.}
  \label{tab:agora_test}
\end{table*}

\subsection{Visualization on in-the-wild scenes}

In this section, we show visualization results of two types of CLIFF estimators with the same HRNet-W48 backbone~\cite{hrnet}. One CLIFF is trained on the BEDLAM dataset (we use the official weights released by authors), annother is trained on both BEDLAM and HumanWild datasets. As we can see from~\cref{fig:hps_res_vis}, our finetuned CLIFF estimator has more accurate pose and shape estimation results in diverse scenes, demonstrating the effectiveness of our data to help existing models have better generalization performance.

\section{Limitations and Future Work}

\begin{figure}[H]
    \centering
    \includegraphics[width=0.95\linewidth]{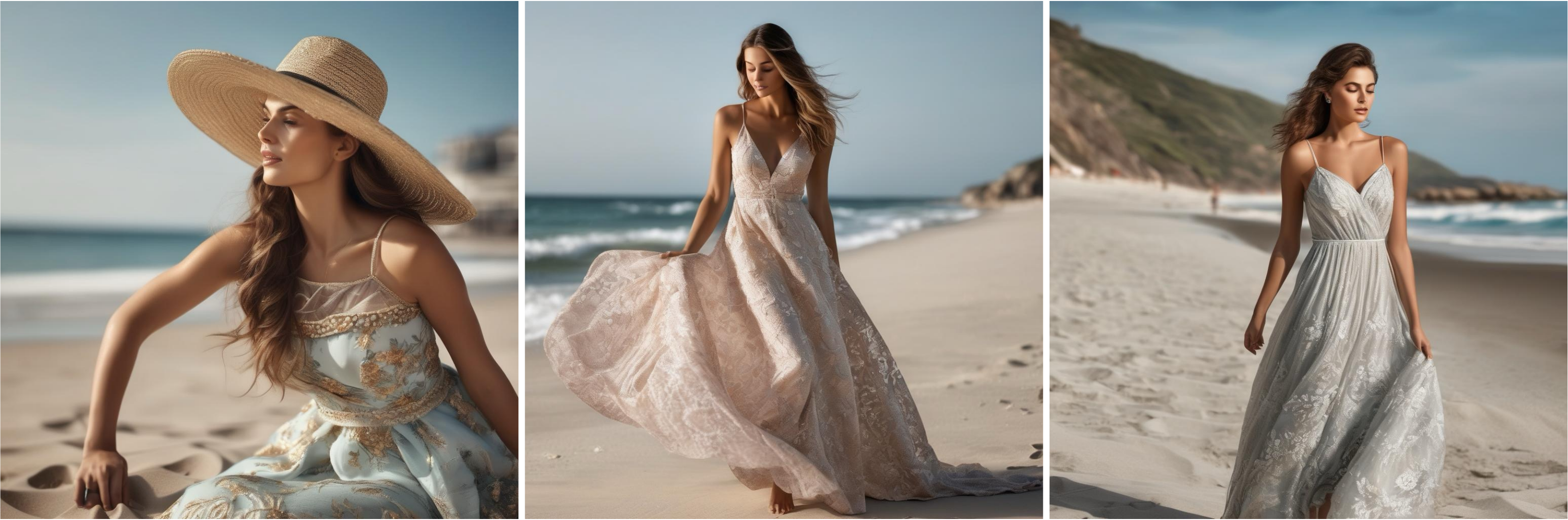}
    \caption{Images generated directly from SDXL~\cite{sdxl}. Text prompt: a beautiful woman wearing a loose dress on the beach.}
    \label{fig: sdxl}
\end{figure}

\noindent\textbf{Limitations in photo-realism.} \added[id=R3]{We observe that our fine-tuned ControlNet model can generate images that are either blurry or lack photorealism. This issue stems from that the underlying SDXL base model possesses limited capacity for photorealistic generation. As shown in~\cref{fig: sdxl}, the SDXL base model tends to generate images with a blurred background. Given recent advancements, such as the FLUX~\cite{flux2024} model, which demonstrate more promising results in synthesizing high-resolution and realistic human imagery, future efforts should be directed toward scaling up controllable generative models to achieve greater photorealism.}

\noindent\textbf{Limitations in multi-Human synthesis.} \added[id=R3]{The current pipeline effectively generates high-quality images containing one to three human subjects. When the number of individuals increases, their scale within the image decreases, leading to poor alignment with the inputs of the guiding depth, keypointmap. Consequently, the resulting data is unsuitable for training human pose and shape models. Future research could leverage recent generative image editing models, such as Bagel~\cite{deng2025bagel} and Flux Kontext~\cite{labs2025flux1kontextflowmatching}, to translate our computer-rendered datasets into more photorealistic domains.}

\section{Discussion and Conclusions}
\label{sec:conclusion}

Based on the experiments, we can answer the question ``Is synthetic data generated by generative models complementary to CG-rendered data for the 3D HPS task?'' Our results strongly suggest that through the integration of 3D human priors with pretrained diffusion models, it becomes feasible to produce high-fidelity pseudo labels encompassing a wide array of real-world scenarios.
As the landscape sees the emergence of increasingly large generative models, there arises a promising prospect for the expansion of diverse 3D human training datasets without intricate mocap systems and CG pipelines.
We hope that our endeavors could pave the way for leveraging generative models to generate high-quality datasets, which are conducive to enhancing the efficacy of 3D human perception tasks.

Our pipeline holds significant potential for various related tasks where acquiring high-quality data pairs is difficult, such as 3D animal pose estimation and 3D reconstruction of human-object or human-human interactions. Addressing these challenges will require further advancements in our methodology, which we will explore in future research.

\clearpage
\newpage

\begin{IEEEbiography}[{\includegraphics[width=1in,height=1.25in,clip,keepaspectratio]{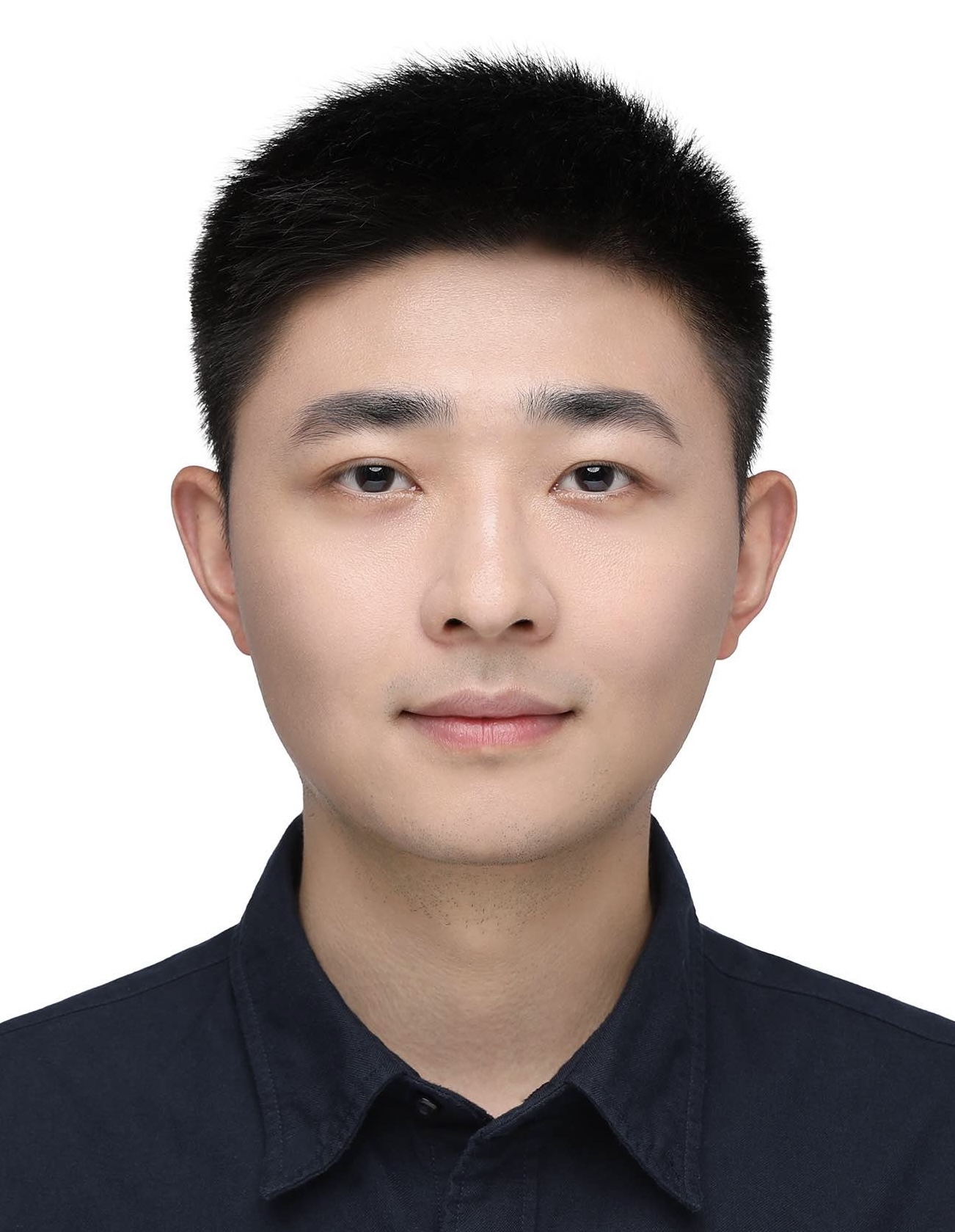}}]{Yongtao Ge}
  Yongtao Ge is currently a PhD student at the University of Adelaide. He has published papers on top-tier conferences, including SIGGRAPH, ICCV, ECCV, ICLR and AAAI. His research interests mainly lies in the intersection of computer vision and computer graphics, including human-centered digitization and world models.
\end{IEEEbiography}

\begin{IEEEbiography}[{\includegraphics[width=1in,height=1.25in,clip,keepaspectratio]{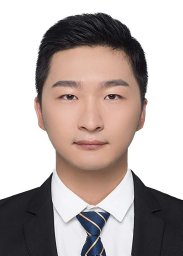}}]{Wenjia Wang}
  is currently pursuing his PhD degree in The Univeristy of Hongkong, advised by Prof.Taku Komura and Prof. Wenping Wang. Prior to that, he worked in computer vision in Sensetime and Tencent, from 2019 to 2022. He has published papers on top-tier conferences, including CVPR, ECCV, ICCV, and IJCAI.
\end{IEEEbiography}

\begin{IEEEbiography}[{\includegraphics[width=1in,height=1.25in,clip,keepaspectratio]{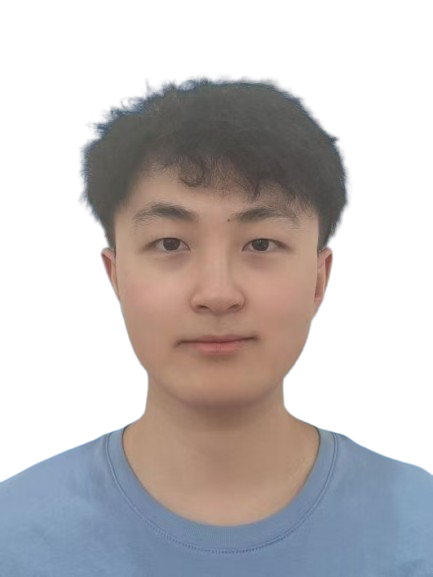}}]{Yongfan Chen}
    was a master student at Zhejiang University, under the supervision of Prof. Chunhua Shen. Prior to this, he earned his bachelor's degree in Computer Science and Technology from Huazhong University of Science and Technology. His current research primarily focuses on computer vision, pose estimation and generative models.
\end{IEEEbiography}

\begin{IEEEbiography}[{\includegraphics[width=1in,height=1.25in,clip,keepaspectratio]{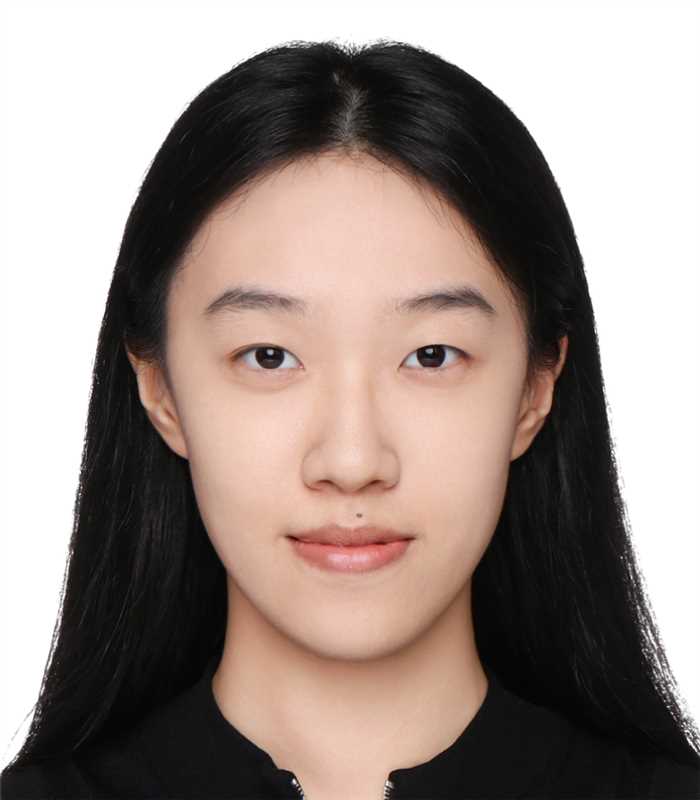}}]{Fanzhou Wang}
   is currently an Algorithm Researcher at SenseTime Group Inc. Prior to that, she received her M.S. degree and B.E. degree from Southeast University in 2023 and 2020. Her research interests include synthetic data and computer graphics.
\end{IEEEbiography}

\begin{IEEEbiography}[{\includegraphics[width=1in,height=1.25in,clip,keepaspectratio]{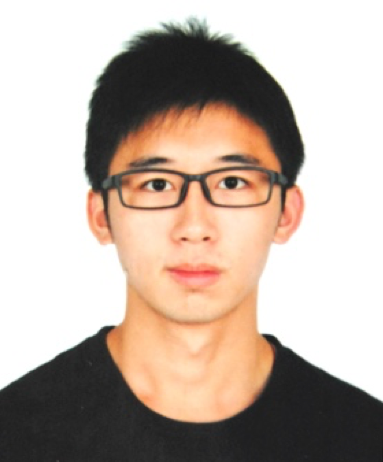}}]{Lei Yang}
  is currently a Research Director at SenseTime Group Inc., under the supervision of Prof. Xiaogang Wang. Prior to that, Lei received his PhD degree from the Chinese University of Hong Kong in 2020, advised by Prof. Dahua Lin and Prof. Xiaoou Tang. Before that, Lei obtained his B.E. degree from Tsinghua University in 2015. He has published over 10 papers on top conferences in relevant fields, including CVPR, ICCV, ECCV, AAAI, SIGGRAPH and RSS.
\end{IEEEbiography}

\begin{IEEEbiography}
[{\includegraphics[width=1in,height=1.25in,clip,keepaspectratio]{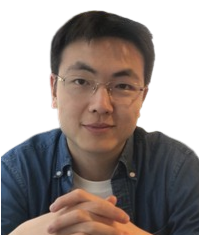}}]{Hao Chen} is an Assistant Professor of Computer Science at Zhejiang University. Prior to that, he worked as a researcher at Noah's Ark Lab at Huawei. He got his PhD degree from University of Adelaide.  He obtained his undergraduate and master degrees from Zhejiang University.
\end{IEEEbiography}

\begin{IEEEbiographynophoto}
{Chunhua Shen}
   is a Chair Professor of Computer Science at Zhejiang University and Zhejiang University of Technology. 
   He was with Amazon Australia, and The University of Adelaide, Australia. His research interest is Computer Vision and Machine Learning. 
\end{IEEEbiographynophoto}
\enlargethispage{-11.5cm}

\clearpage

{\small
\bibliographystyle{unsrt}
\bibliography{main}
}
\newpage

\begin{appendices}

\renewcommand{\thesection}{\Alph{section}}
\section{}
\subsection{Implement Details for Training ControlNet}
\label{sec:supp1}
HumanWild employs a customized multi-condition ControlNet to generate human images and corresponding annotations. In the preliminary phase of our research, we adopt the off-the-shelf depth-to-image ControlNet, and keypoint-to-image ControlNet, which are trained on the LAION5B dataset without filtering low-quality images. Experiments show that the pre-trained models face challenges when it comes to generating human images
with hard poses. To resolve this issue, we collect high-quality human images from multiple datasets and then conduct finetuning on the collected datasets for better generation performance.
We train the customized ControlNet with with AdamW~\cite{adam} in 1e$-5$ learning rate, and 0.01 weight decay for $80,000$ iterations on 8 A100 GPUs. For training efficiency, we utilize Fully Sharded Data Parallel (FSDP) with ZeRO Stage 2, gradient checkpointing, and mixed-precision training

\begin{figure}[ht]
    \centering
    \includegraphics[width=0.48\textwidth]{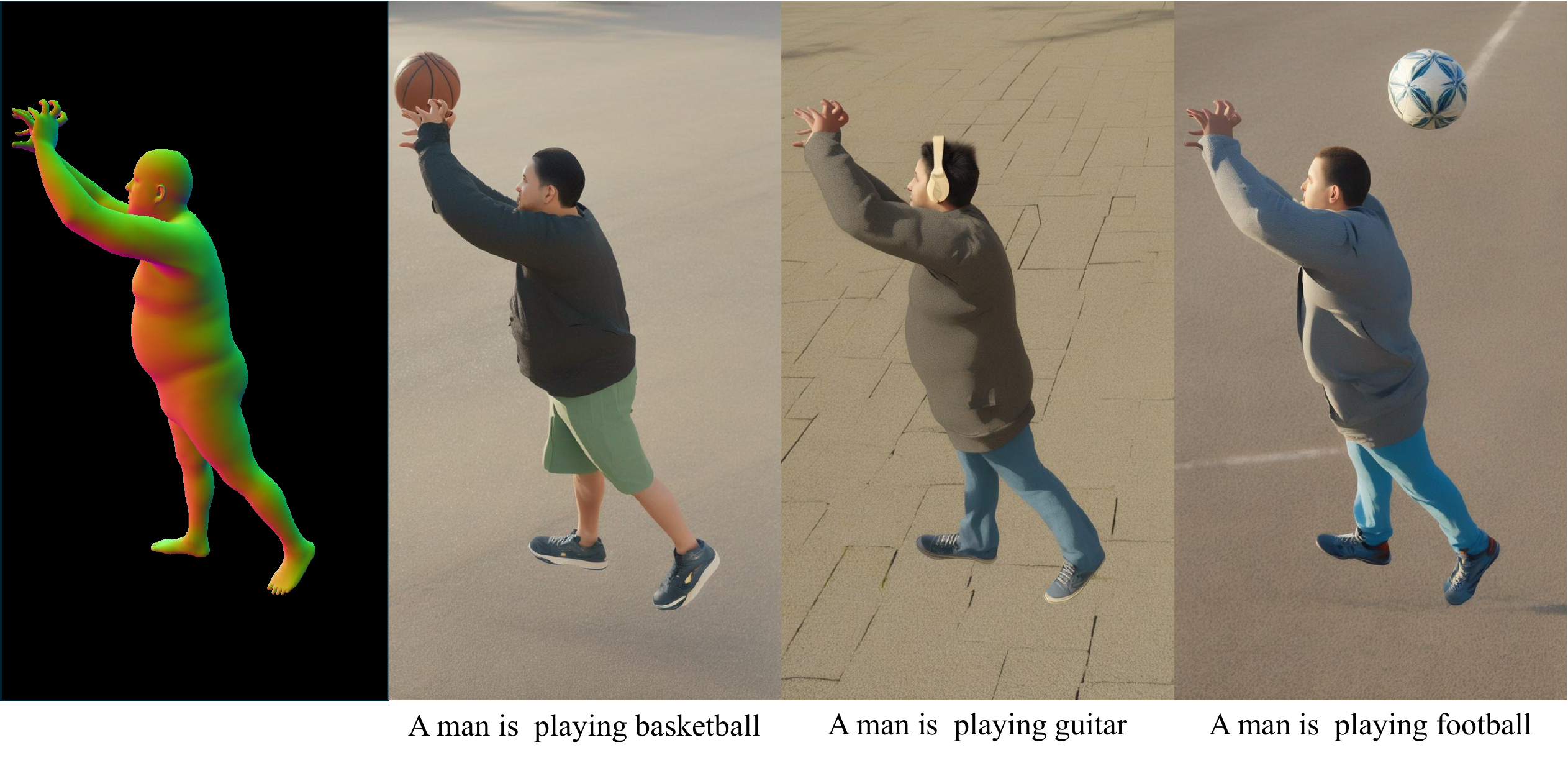}
    \caption{Utilizing identical surface normal maps and control scale factor of the surface normal as input, we change the input text prompts.}
    \label{fig:text_prompt_failure_1}
\end{figure}

\begin{figure}[ht]
    \centering
    \includegraphics[width=0.48\textwidth]{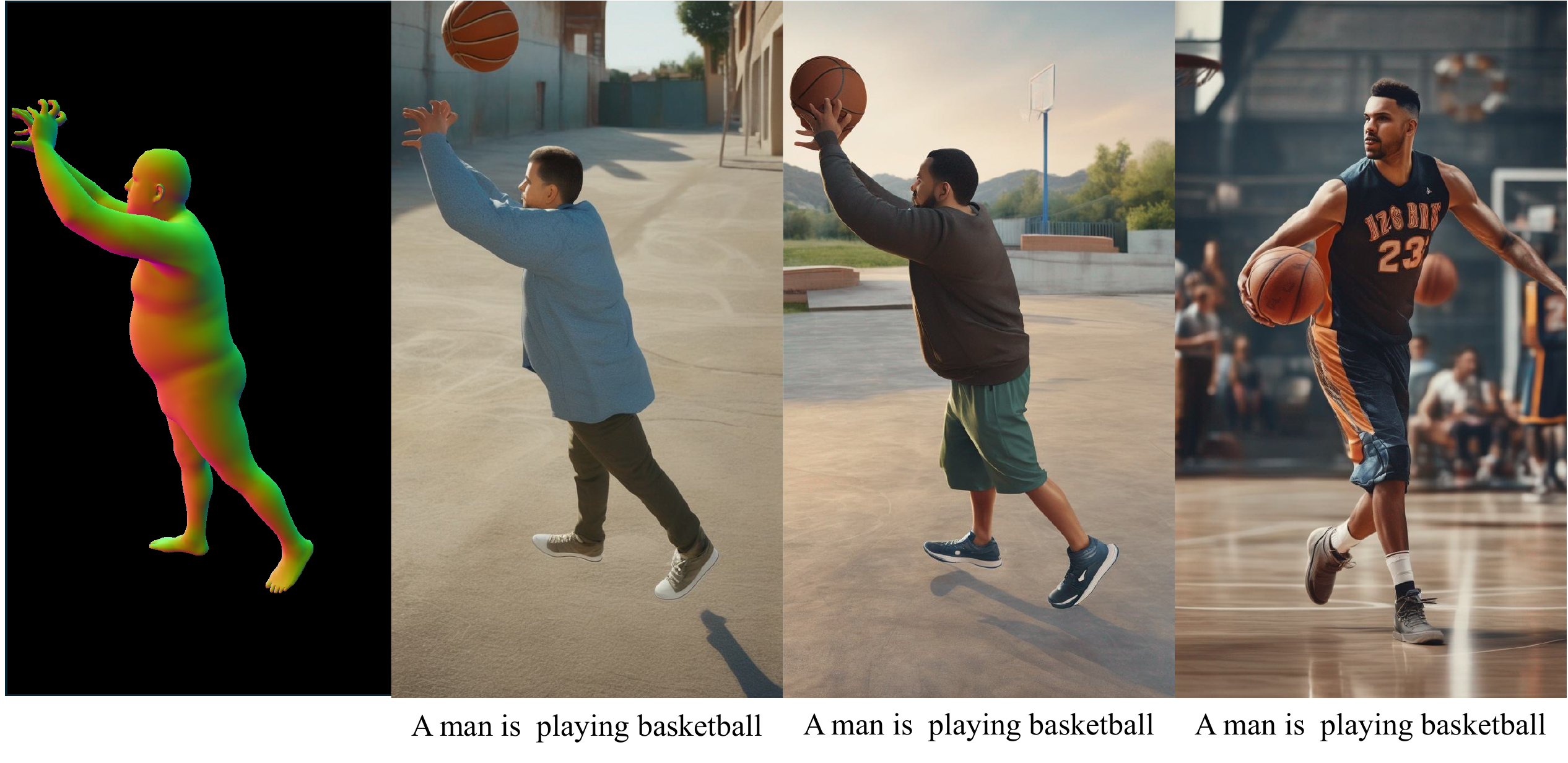}
    \caption{Utilizing identical surface normal maps and text prompts as input, we manipulate the control scale factor of the surface normal map across values of $0.75$, $0.5$, and $0.25$, respectively. }
    \label{fig:text_prompt_failure_2}
\end{figure}

\section{}
\subsection{Failure Cases of the Off-the-shelf ControlNet}
\begin{figure}[t]
	\begin{center}
		\includegraphics[width=\linewidth]{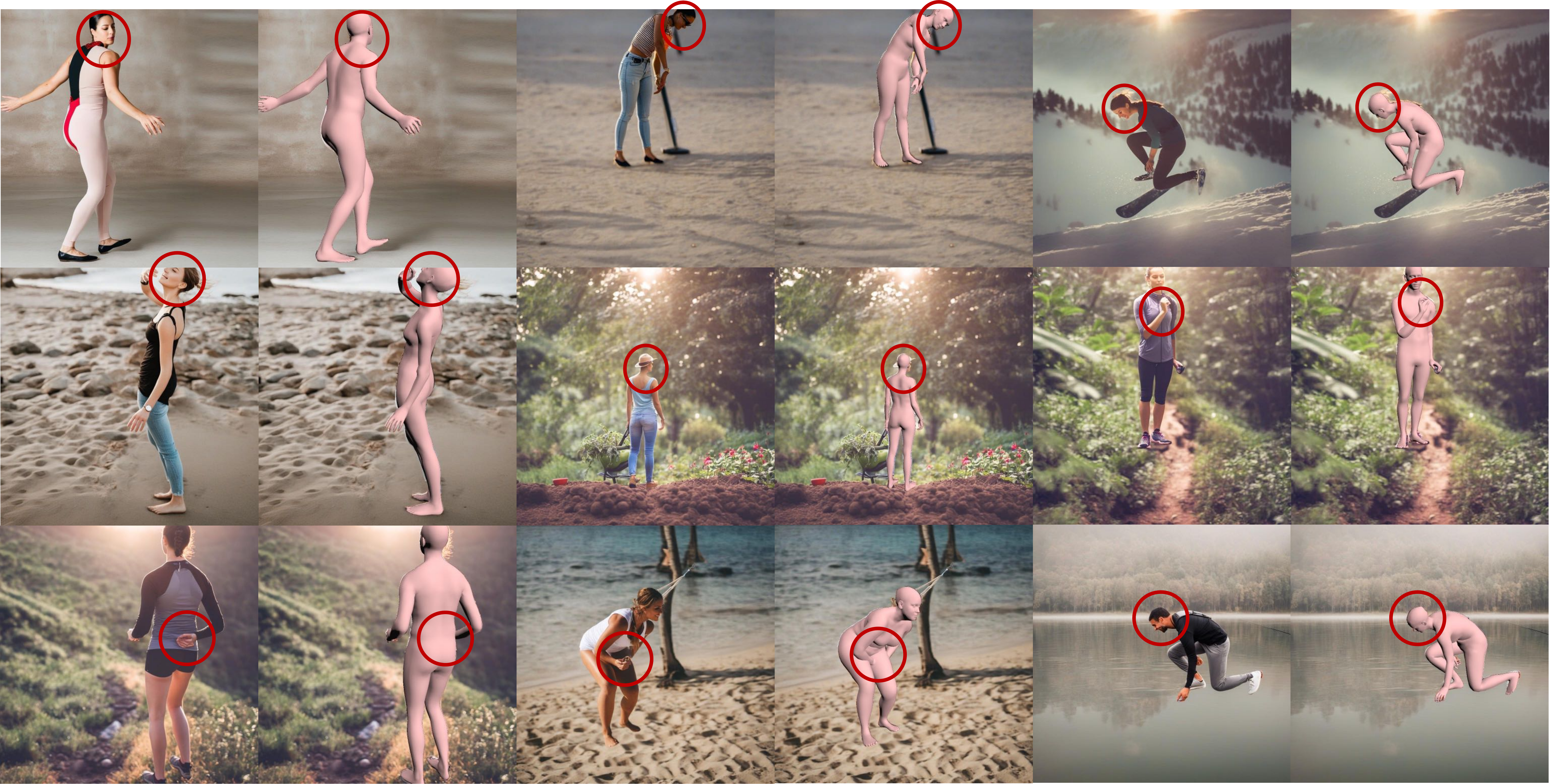}
	\end{center}
	\vspace{-0.0cm}
	\caption{Failure cases of the data pairs generated by general ControlNet. See highlighted regions with red circles.}

	\label{fig: failure_cases}
\end{figure}

In~\cref{fig: failure_cases}, we show some failure cases of the data pairs generated by off-the-shelf ControlNet trained on general datasets. The inconsistency of the image and 3D mesh would affect the performance of the 3D human pose estimation. Thus, it is necessary to retrain a customized multi-condition ControlNet with the human-centeric dataset collected in this work.

\subsection{More Visualization of Generated Data Sample}
\begin{figure*}[tp]
    \centering
    \includegraphics[width=0.98\linewidth]{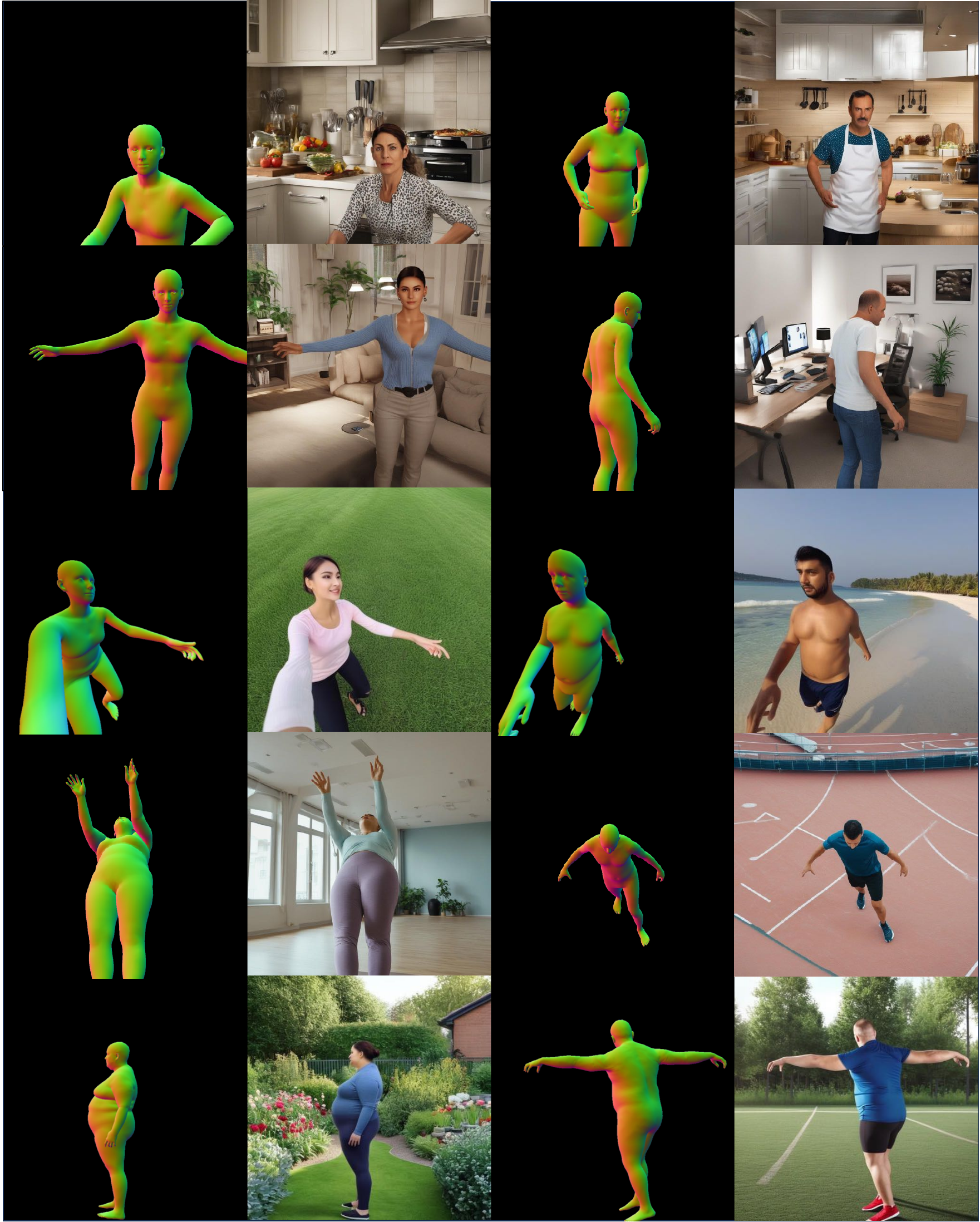}
    \caption{More Data sample visualization of HumanWild. For each data sample, the left side is the normal image rendered from SMPL-X model, the right side is the image generated by HumanWild pipeline. The first two rows demonstrate the indoor activities. The third row shows the generated images with huge distortion. The fourth row describes the diverse camera views. The fifth row shows overweight human bodies. }
\label{fig:gt_vis}
\end{figure*}

We show more data sample visualization with diverse scenes and body shapes in ~\cref{fig:gt_vis}.

\begin{figure*}[ht]
	\begin{center}
		\includegraphics[width=0.95\linewidth]{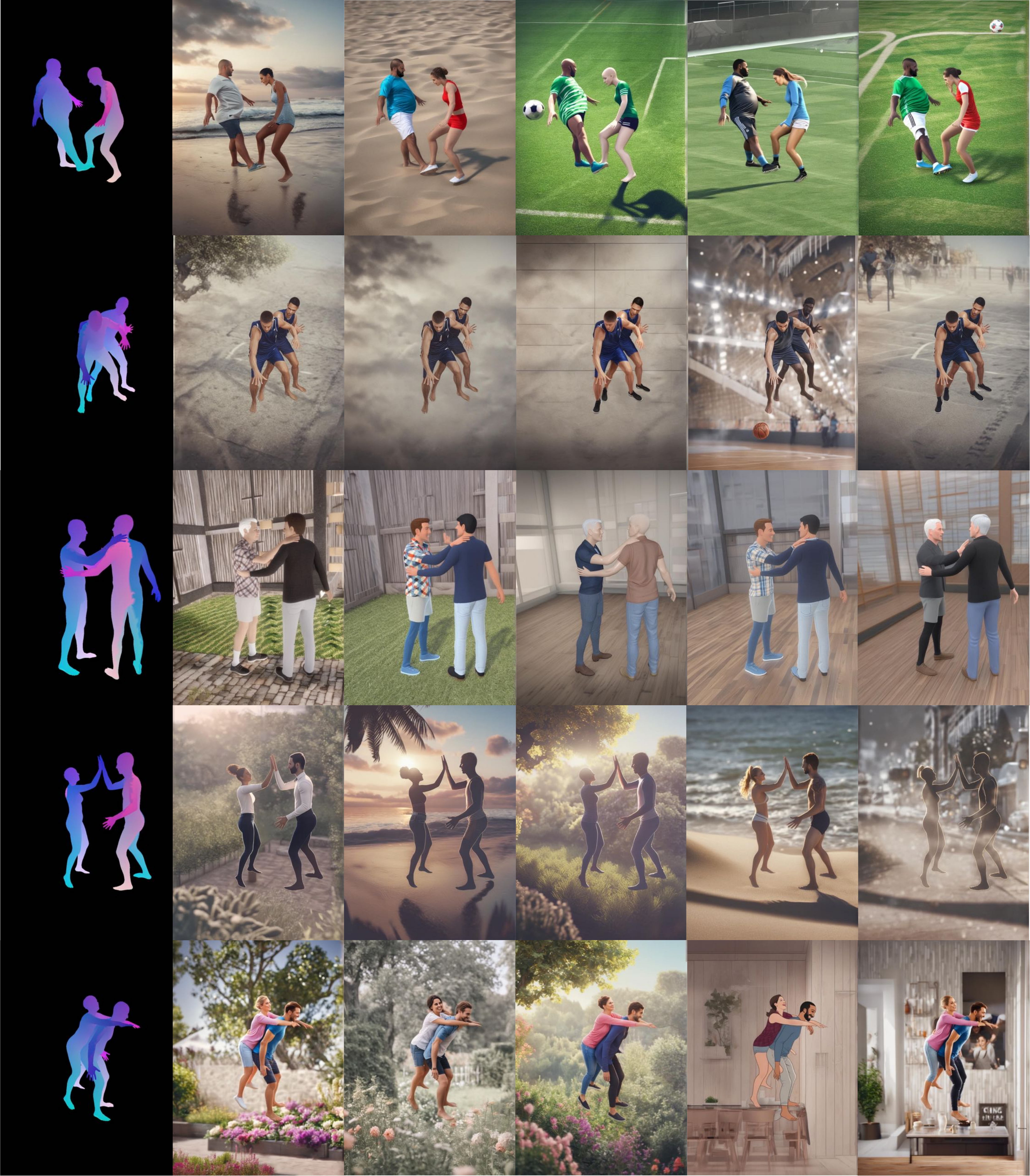}
	\end{center}
	\vspace{-0.0cm}
	\caption{Visualization of Human interaction. The SMPL interaction annotations
	are sampled from the Hi4D~\cite{yin2023hi4d} dataset.}

	\label{fig: interaction}
\end{figure*}

We show visualization results on human interactions in \cref{fig: interaction}, which demonstrates that our pipeline can generate well-aligned image-annotation pairs where people are with close interactions. The generated data pairs are of great value in enhancing existing human interaction datasets collected in the studio environment. (\eg, Hi4D~\cite{yin2023hi4d} and CHI3D~\cite{Fieraru_2020_CVPR})

\subsection{Text Prompt Examples Generated by LLM}

In~\cref{tab: prompts}, we show some text prompt examples, which are generated
by ChatGPT~\cite{openai2023gpt4} with diverse human actions and scenes.

\begin{table}[ht]
	\caption{Text prompt examples.}
	\label{tab: prompts} \scriptsize
	\centering
	\resizebox{0.8\columnwidth}{!}{
	\begin{tabular}{l|l|l}
		\toprule \multicolumn{1}{l|}{gender} & \multicolumn{1}{l|}{action} & \multicolumn{1}{l}{environment} \\
		\midrule a man                       & playing soccer              & at the park                     \\
		a woman                              & swimming                    & in the pool                     \\
		a man                                & shopping                    & at the mall                     \\
		a woman                              & running                     & in the park                     \\
		a man                                & studying                    & at the library                  \\
		a man                                & working                     & at the office                   \\
		a man                                & chatting                    & at a cafe                       \\
		\bottomrule
	\end{tabular}
	}

\end{table}

\subsection{Misalignment Analysis of Complicated Prompts and Spatial Conditions}

We present instances of failure where text prompts conflict with the spatial condition, \eg, surface normal. Specifically, when the text prompt suggests an action that deviates from the surface normal map, the resulting images often fail to adhere to the text prompt, particularly when the control factor of the surface normal map approaches $1.0$.
As shown in \cref{fig:text_prompt_failure_1}, despite employing the same input surface normal map and a control factor of $0.95$, the generated images exhibit similar foreground human identities but differ in background elements corresponding to distinct text prompts.
\cref{fig:text_prompt_failure_2} illustrates that when the control factor of the surface normal is extremely low, the generated images may disregard the surface normal map condition.

\clearpage

\end{appendices}

\end{document}